%% file: main.tex
\documentclass[journal]{IEEEtran}
\usepackage{amsmath,amsfonts}
\usepackage{epsfig, amssymb}
\usepackage{hyperref}
\usepackage{algorithmic}
\usepackage{algorithm}
\usepackage{array}
\usepackage[caption=false,font=normalsize,labelfont=sf,textfont=sf]{subfig}
\usepackage{textcomp}
\usepackage{stfloats}
\usepackage{url}
\usepackage{verbatim}
\usepackage{graphicx}
\usepackage{cite}
\usepackage{mathtools}
\usepackage{bbold}
\usepackage{ntheorem}
\usepackage{tikz}
\usepackage{pgfplots}
\pgfplotsset{compat=1.18}
\input{math_command.tex}
\theoremstyle{break}
 
\newtheorem{theorem}{Theorem}
\newtheorem{lemma}[theorem]{Lemma} 
\newtheorem{proposition}[theorem]{Proposition} 
\newtheorem{remark}[theorem]{Remark}

\newtheorem{definition}[theorem]{Definition}

\newtheorem{assumption}[theorem]{Assumption}

\theoremstyle{nonumberbreak}
\newtheorem{assumptionnn}{Assumption 0}

\begin{document}

\title{A Large Dimensional Analysis of Multi-task Semi-Supervised Learning}

\author{Victor Léger,~\IEEEmembership{Student member,~IEEE}, Romain Couillet,~\IEEEmembership{Senior member,~IEEE}}





\maketitle

\begin{abstract}
This article conducts a large dimensional study of a simple yet quite versatile classification model, encompassing at once multi-task and semi-supervised learning, and taking into account uncertain labeling. Using tools from random matrix theory, we characterize the asymptotics of some key functionals, which allows us on the one hand to predict the performances of the algorithm, and on the other hand to reveal some counter-intuitive guidance on how to use it efficiently. The model, powerful enough to provide good performance guarantees, is also straightforward enough to provide strong insights into its behavior.
\end{abstract}


\section{Introduction}
\label{sec:intro}

If the multiple achievements of deep neural networks (DNNs) in various tasks during the last decade are now a well-known fact, we also know that these achievements have been strongly reliant on increases in computing power \cite{thompson_computational_2022}. 
This escalation implies skyrocketing energy requirements, but another fondamental consequence relates to tractability, and even technical accessibility, which are, in the case of DNNs, out of reach to their own designers (let alone to their users).

As an alternative viewpoint, following the notion of \emph{conviviality of tools} developed by the critical analyst Ivan Illich \cite{illich1973}, the present article is part of a renewed endeavour for techniques and tools which simultaneously \emph{answer an effective need} (rather than create a new need), \emph{are tractable and easily apprehensible by their user}, and \emph{are flexible and robust enough not to require constant adaptations and additional scientific complexification}. 

That is, through the present article, we aim to open the path to all-encompassing flexible tools for classification problems, buttressed on elementary machine learning notions and rather basic mathematical tools. Specifically, we propose here a classification method that addresses a wide variety of real-life conditions, such as the possibility for only part of the input data to be prelabeled (thereby encompassing supervised, semi-supervised and unsupervised learning at once), the possibility for erroneous or uncertain labels (here considering biases or controversial data labeling), the possibility for strong imbalances in prelabeled classes of data (a very classical issue when some key classes are rarely observed), etc. As such, our base model combines what are usually considered as distinct learning frameworks (multi-task learning, semi-supervised learning, uncertain labeling) under a common umbrella.

Specifically, we develop a simple and quite technically accessible method which perfoms linear classification in a multi-task and semi-supervised framework, in a basic two-class setting (the extension to a multiclass setting is briefly discussed in Section~\ref{sec:multiclass}). Considering Gaussian mixtures as data model, we leverage random matrix theory (RMT) to characterize the asymptotics of key functionals, such as the \emph{decision function}, which enables classification by attributing a score to each new sample. In contrast to classical statistics, we consider high dimensional data (\ie the dimension $p$ and the number of data $n$ are of the same order of magnitude -- which is more a fact than an assumption in modern data). 

The relative simplicity of our model, along with the strength of RMT, make the proposed method quite powerful as it is in particular able to \emph{predict its own performances}. 
In a nutshell, our theoretical analysis allows us to:
\begin{itemize}
    \item Tune the hyperparameters of the model without running the algorithm, therefore avoiding any cross-validation (costly in data and computational resources).
    \item Choose wisely (and automatically) the values of labels (rather than the standard $\pm 1$ values) to discriminate more efficiently the data, and tackle the problems induced by unbalanced classes.
    \item Understand better and avoid the classical problem of \textit{negative transfer}.
    \item Estimate how good the algorithm is in comparison with optimal bounds from information theory.
\end{itemize}

Both multi-task learning and semi-supervised learning have already been studied with RMT tools, usually separately though. Specifically, \cite{mai2021consistent} provides a theoretical analysis (in a single-task framework) of the graph-based semi-supervised scheme we use in this article \cite{zhou2003learning,avrachenkov_generalized_2012}. In particular, it reveals a fundamental flaw of graph-based algorithms, but most importantly, it proposes a simple way to tackle it. 
Indeed, many algorithms appear to be unable to learn effectively from unlabeled data, even though it is a fundamental aspect of semi-supervised learning \cite{shahshahani1994,Cozman2006RisksOS,10.7551/mitpress/9780262033589.001.0001,BenDavid2008DoesUD}.

Meanwhile, \cite{tiomoko2020large} provides a similar analysis of a well-known multi-task framework \cite{evgeniou2004}, in a fully supervised setting. In particular, it introduces (for the first time, as far as we know) the counter-intuitive idea of choosing the values of labels differently from the standard $\pm 1$. This idea, which is one of the keys to our present algorithm, appears to be an easy and intuitive way to tackle the issue of negative transfer.

The remainder of the article is organized as follows. Section~\ref{sec:model} introduces the assumptions made, the problem formulation and its solution. Section~\ref{sec:results} presents the main theoretical results of the article, upon which our algorithm is built. Section~\ref{sec:uncertain} extends these results to the case of uncertain labeling. Section~\ref{sec:limitations} sheds light on some limitations of our method. Finally, Section~\ref{sec:exp} presents simulations on both synthetic data and real datasets, to validate and further analyse the results of Section~\ref{sec:results}.

In what follows, matrices are represented by bold uppercase letters (\ie matrix $\mA$). Vectors are denoted in bold lowercase letters (\ie vector $\vv$) and scalars without bold letters (\ie variable $a$). 
``tasks'' are denoted by an index $t$ and the number of tasks will be referred as $T$. Quantities with superscript $t$ are belonging to Task $t$.
The quantities associated to labeled and unlabeled data are respectively denoted by a subscript $\ell$ and $u$. As such the notations $n_\ell^{t}$ and $n_u^{t}$ denote respectively the number of labeled and unlabeled data of Task $t$. The subscript $j$ is used to specify that a quantity is specific to the Class $j$.

\section{Model and assumptions}
\label{sec:model}
We focus from now on the two-class setting, the multi-class setting is briefly discussed in Section \ref{sec:multiclass}.
Let $\mX\in \R^{p\times n}$ be a collection of $n$ independent data vectors of dimension $p$. The data are divided into $T$ subsets attached to individual ``task''. Specifically, letting $\mX=[\mX^{1},\ldots,\mX^{T}]\in\mathbb{R}^{p\times n}$, Task $t$ is a semi-supervised binary classification task with training samples $\mX^{t}=[\mX_\ell^t,\mX_u^t]\in\R^{p\times n^{t}}$ which consists of a set of $n_\ell^{t}$ labeled data samples $\mX_\ell^{t} = \{\vx_{i}^{t}\}_{i=1}^{n_\ell^{t}}$ and a set of $n_u^{t}$ unlabeled data points $\mX_u^{t} = \{\vx_{i}^{t}\}_{i=n_\ell^{t}+1}^{n^{t}}$.
To each labeled data $\vx_{i}^{t}$ from Task $t$ is attached a label $y_{i}^{t}$ and the goal is to predict the labels for the unlabeled data $\mX_u^{t}$.
In this paper, we will divide the $T$ tasks in:
\begin{itemize}
    \item One ``target task'' which is the one task we actually want to perform.
    \item $T-1$ ``source tasks'' which will help us to perform the ``target task''.
\end{itemize}
In a sense, our setting is close to transfer-learning because we aim to perform a single classification task. However, as the calculus done in the following is true for any choice of target task, we can easily perform sequentially each one of the $T$ tasks with the help of the other, and therefore perform multi-task learning.

\begin{assumptionnn}[On the data distribution]
 \label{ass:data_distribution_1}
The columns of the data matrix $\mX$ are independent Gaussian random variables. Specifically, the data samples $\left(\vx_1^{t},\dots,\vx^{t}_{n^{t}}\right)$ from Task $t$ are i.i.d.\@ observations such that $\vx_i^{t} \in \mathcal{C}_j^{t} \Leftrightarrow \vx_i^{t} \sim \mathcal{N}(\vmu_j^{t},\mI_p)$ where $\mathcal{C}_j^{t}$ denotes the Class $j$ of Task $t$.
\end{assumptionnn}

Even though this assumption is intuitive and straightforward, one could argue that it is unrealistic. Indeed, natural data is  not Gaussian, and most importantly, requiring independent entries for each observation is very constraining. Concentrated vectors, described for the first time in \cite{ledoux2001} and integrated in RMT by \cite{El_Karoui_2009}, allow us to relax the previous assumption, while preserving its core idea.

\begin{assumption}[On the data distribution]
\label{ass:data_distribution_2}
The columns of the data matrix $\mX$ are independent concentrated vectors with isotropic covariance. Specifically, the data samples $\left(\vx_1^{t},\dots,\vx^{t}_{n^{t}}\right)$ from Task $t$ are i.i.d.\@ observations such that: \\
There exists $C,c>0$ such that for any $1$-Lipschitz function $\Phi : \mathbb{R}^p \to \mathbb{R}$, we have:
\begin{equation*}
\forall u>0, \quad \vx_i^{t} \in \mathcal{C}_j^{t} \Rightarrow \mathbb{P}(|\Phi(\vx_i^{t})-m_j^t(\Phi)|\geq u) \leq Ce^{-(\frac{u}{c})^2}
\end{equation*}
where $\mathcal{C}_j^{t}$ denotes the Class $j$ of Task $t$, and $m_j^t(\Phi) = \mathbb{E}\left[\Phi(\vx)|\vx \in \mathcal{C}_j^{t}\right]$. \\
Under these assumtions, we can define
\begin{align*}
    \vmu_j^t &= \mathbb{E}[\vx|\vx \in \mathcal{C}_j^{t}] \\
    \mSigma_j^t &= \Cov[\vx|\vx \in \mathcal{C}_j^{t}]
\end{align*}
and we further impose that $\forall j,t, \mSigma_j^t = \mI_p$.
\end{assumption}

This second assumption encompasses the first one, and applies to a broader class of distributions. In particular, images produced by Generative Adversarial Networks (GANs) are concentrated vectors \cite{Seddik2020RandomMT}. As GANs are known to produce very convincing fake images and text, this assumption allows to consider more realistic datasets. All the theoretical results discussed in the remainder of the article are obtained using Assumption \ref{ass:data_distribution_2}. The assumption $\mSigma_j^t = \mI_p$ is still restrictive, and should not be seen as an imperious condition for our algorithm to work, but rather as a convenient assumption to perform our statistical analysis. As such, the experiments of Section \ref{sec:real_data} shows that the algorithm performs well on concentrated vectors for which the covariance is not necessarily isotropic.

As discussed earlier, we consider a large dimensional setting, where the dimension of the data and the number of data have the same order of magnitude. Moreover, to take into account the potential consequences of class unbalances, our statistical analysis will make use of the proportion of data in each class. To this end, we make the following assumption.

\begin{assumption}[Growth Rate]
\label{ass:growth_rate}
As $n\to \infty$,
\begin{itemize}
    \item $p/n \to c>0$.
    \item $n_{j}^{t}/n\to \rho_{j}^{t}>0$ ($n_{j}^{t}$ the number of data in $\mathcal{C}_j^t$).
    \item $n^{t}/n\to \rho^{t}>0$.
    \item $n_{\ell j}^{t}/n_{j}^{t} \to \eta_{j}^{t}>0$ ($n_{\ell j}^{t}$ the number of labeled data in $\mathcal{C}_j^t$).
    \item $n_{\ell}^{t}/n^{t} \to \eta^{t}>0$.
\end{itemize}
Furthermore, we also consider the $2T$-dimensional vector $\vrho=\left(\rho_{1}^{1},\rho_{2}^{1},\rho_{1}^{2},\dots,\rho_{2}^{T}\right)^\trans$, the $T$-dimensional vector $\bar{\vrho} = \left(\rho^{1},\dots,\rho^{T}\right)^\trans$, and similarly the vectors $\veta$ and $\bar{\veta}$.
\end{assumption}

Graph-based approaches, such as the Laplacian regularization method, are widely use to perform semi-supervised learning (in a single-task setting) \cite{joachims_transductive_2003,zhou2003learning,zhu_semi-supervised_2003}. The idea consists in propagating the effective information of labeled data to unlabeled data, following the natural assumption that similar data points should have similar labels. This similarity between data points $\vx_i$ and $\vx_{i'}$ is measured by the quantity $\omega_{ii'}=h\left(\frac{1}{p}\langle \vx_i,\vx_{i'}\rangle\right)$ with $h$ an increasing function ($h(x)=x$ in our case), so that similar data vectors $\vx_i$ and $\vx_{i'}$ are connected with a large weight. From there, graph-based learning algorithms estimate the class of each node $\vx_i$ through a class attachment ``score'' $f_i$, by solving the following optimization problem:
\begin{align}
\label{eq:opti_pb_st}
    &\min\limits_{\vf}\sum\limits_{i,i'=1}^{n}\omega_{ii'}\left(f_i-f_{i'}\right)^2\\
    &\textit{such that} ~  f_i=y_i ~ \forall ~ 1\leq i\leq n_{\ell}.\nonumber
\end{align}

The term $f_i=y_i$ is the fitting constraint, which imposes that the score of labeled data should match the initial label assignment. This constraint is sometimes relaxed by adding a term $2\alpha\|\vf_\ell-\vy_\ell\|^2$ to the minimization problem, with $\alpha>0$ \cite{avrachenkov_generalized_2012}, where $\vf_\ell$ and $\vy_\ell$ respectively denotes the vectors $\vf$ and $\vy$ restricted to labeled data.

Our multi-task semi-supervised algorithm follows the same idea, yet with the addition of a hyperparameter matrix $\mLambda = \{\Lambda^{tt'}\}_{t,t'=1}^{T}$ which filters how much each task should be related to each other. The optimisation becomes: 

\begin{align}
\label{eq:opti_pb}
    &\min\limits_{\vf^{1},\ldots,\vf^{T}} \sum\limits_{t,t'=1}^T \Lambda^{tt'}\sum\limits_{i=1}^{n^{t}}\sum\limits_{i'=1}^{n^{t'}}\omega_{ii'}^{tt'}\left(f_i^{t}-f_{i'}^{t'}\right)^2\\
    &\textit{such that} ~  f_i^{t}=y_{i}^{t} ~ \forall ~ 1\leq i\leq n_{\ell}^{t} ~ \textit{and} ~ 1\leq t\leq T.\nonumber
\end{align}
and the weights $\omega_{ii'}^{tt'}$ are now
\begin{equation*}
    \omega_{ii'}^{tt'}=\frac{1}{Tp}\langle \vx_i^{t},\vx_{i'}^{t'}\rangle
\end{equation*}

The classical Laplacian regularization algorithm associated to \eqref{eq:opti_pb_st} has been studied in depth in \cite{mai2018random} in the single-task setting. There, the authors showed the fundamental importance to ``center'' the weight matrix $\mW = \{\omega_{ii'}\}_{i,i'=1}^{n}$. This centering approach corrects an important bias in the regularized Laplacian which completely annihilates the use of unlabeled data in a large dimensional setting. A significant performance increase was reported, both in theory and in practice in \cite{mai2021consistent} when this basic, yet counter-intuitive, correction is accounted for. The same phenomenon evidently arises in the multi-task extension and we propose consequently the same \textit{task-wise} pre-centering in the context of multi-task learning, that is we update the weight matrices $(\mW^{tt'})_{t,t'}$ as:
\begin{equation}
\label{eq:centering}
     \hat{\mW}^{tt'}=\mP^{t}\mW^{tt'}\mP^{t'}
 \end{equation}
with $\mP^{t}= \left(\mI_{n^t}-\frac{1}{n^t}\mathbb{1}_{n^t}\mathbb{1}_{n^t}^\trans\right)$ the centering projector. This is equivalent to replace the data matrix $\mX^t$ by its centered version $\mathring{\mX}^t=\mX^t\mP^{t}$. It is worth to note that the centering is performed undifferently on labeled and unlabeled data, as the whole task is concerned and both classes are centered at once. For a matter of readability, $\mX^t$ will denote the centered matrix in the remainder of the article. Similarly, $\vmu_j^t$ will denote the centered mean of $\vx\in\mathcal{C}_j^t$.

However, the optimization problem described in \eqref{eq:opti_pb} is non convex since the entries of the weight matrix $\hat{\mW}$ may take negative values (this must actually be the case as the mean value of the entries of $\hat{\mW}$ is zero). To deal with this problem, we further propose (as in \cite{mai2021consistent}) to constrain the norm of the unlabeled data score vector $\vf_{u}$ (that is, the score vector $\vf$ restricted to unlabeled data) by appending a regularization term $2\alpha\|\vf_u\|^2$ to the previous minimization problem, with $\alpha>0$. We also relax the labeling constraint with the regularization term $2\alpha\|\vf_\ell-\vy_\ell\|^2$ mentionned above.
The equation \eqref{eq:opti_pb} becomes:
\begin{align}
\label{eq:opti_pb_2}
    &\min\limits_{\mathbf{f}^{1},\ldots,\mathbf{f}^{T}} \sum\limits_{t,t'=1}^T \Lambda^{tt'}\sum\limits_{i=1}^{n^{t}}\sum\limits_{i'=1}^{n^{t'}}\hat{\omega}_{ii'}^{tt'}\left(f_i^{t}-f_{i'}^{t'}\right)^2 \\
    &+ 2\alpha \sum\limits_{t=1}^T\left(\sum\limits_{i=1}^{n_\ell^t} (f_i^t-y_i^t)^2 + \sum\limits_{i=n_\ell^t+1}^{n^{t}} (f_i^t)^2\right) \nonumber
\end{align}

This leads, under a more convenient matrix formulation (see detail in Section~\ref{app:optimization} of the appendix), to
\begin{equation}
\label{eq:opti_pb_3}
    \min\limits_{\vf \in \mathbb{R}^{n}} -{\vf}^{\trans} \tilde{\mW} \vf + \alpha\|\vf_\ell-\vy_\ell\|^2 + \alpha\|\vf_u\|^2 
\end{equation}
with $\tilde{\mW}$ the block matrix for which each block is $\tilde{\mW}^{tt'}=\Lambda^{tt'}\hat{\mW}^{tt'},\forall (t,t')\in \{1,\dots,T\}^2$.
If we further use the convention $\vy_u=0$, the problem has an even simpler formulation:
\begin{equation}
    \min\limits_{\vf \in \mathbb{R}^{n}} \alpha\|\vf-\vy\|^2 - {\vf}^{\trans} \tilde{\mW} \vf
\end{equation}

This problem is now convex for all $\alpha > \|\tilde{\mW}\|$. In addition, $\vf$ is the solution to a quadratic optimization problem with linear equality constraints, and can be obtained explicitly (see detail in Section~\ref{app:optimization} of the appendix).
\begin{equation}
\label{eq:solution}
    \vf = \left(\mI_{n} - \frac{\mZ^\trans \mA \mZ}{Tp} \right)^{-1}\vy
\end{equation}
where
\begin{align*}
    \mA&=\tilde{\mLambda}  \otimes \mI_p, ~ \textit{with} ~ \tilde{\mLambda} = \frac{\mLambda}{\alpha} ~ \textit{the hyperparameter matrix}\\
    \mZ&=\sum\limits_{t=1}^T \mE_{tt}^{[T]}\otimes \mX^{t} ~  \textit{the data matrix}\\
\end{align*}

If we further use Woodbury matrix identity, we have
\begin{equation}
    \vf = \vy + \frac{1}{Tp}\mZ^\trans\mA^{\frac 12}\underbrace{\left(\mI_{Tp} - \frac{\mA^{\frac 12} \mZ\mZ^\trans\mA^{\frac 12}}{Tp} \right)^{-1}}_{=\mQ}\mA^{\frac 12}\mZ\vy
\end{equation}
This last solution has the benefit of being explicit and therefore easily tractable. It can be interpreted as being an \textit{a priori} on the scores (the first term), corrected by the data (the second term). As for unlabeled data, there should not be any \textit{a priori}, which justifies the previous choice of $\vy_u=0$.
Focusing on the scores of unlabeled data then leads to:
\begin{equation}
    \vf_u = \frac{1}{Tp}\mZ_u^\trans\mA^{\frac 12}\mQ\mA^{\frac 12}\mZ_\ell\vy_\ell
\end{equation}

To reduce the number of parameters and simplify the model, we choose to set as of now the values of $\mLambda$ as follows:
\begin{equation}
\label{eq:lambda}
    \Lambda^{tt'} = \frac{|\langle \boldsymbol{\mu}_1^t-\boldsymbol{\mu}_2^t,\boldsymbol{\mu}_1^{t'}-\boldsymbol{\mu}_2^{t'} \rangle|}{\|\boldsymbol{\mu}_1^t-\boldsymbol{\mu}_2^t\|\|\boldsymbol{\mu}_1^{t'}-\boldsymbol{\mu}_2^{t'}\|}.
\end{equation}
As expected $\Lambda^{tt'}$ increases if the tasks are more correlated. This quantity is intuitively associated to the alignement between tasks and can be consistently estimated. Most importantly, simulations comparing this choice to an optimization of $\mLambda$ on $[0,1]^{T\times T}$ suggests that this choice is close to optimal.

\section{Main results}
\label{sec:results}

We have an explicit formulation of the score vector $\vf_u$ of unlabeled data. This vector sums up all the information we have to perform the classification. In a classical machine learning setting, the score $f$ of a sample $\vx$ belonging to $\mathcal{C}_1^t$ (resp. $\mathcal{C}_2^t$) is expected to be close to $-1$ (resp. $+1$). Therefore, the classification rule is:
\begin{align}
\label{eq:classif}
    f<\zeta^t &\implies \vx\to\mathcal{C}_1^t \\
    f\geq\zeta^t &\implies \vx\to\mathcal{C}_2^t \nonumber
\end{align}
with $\zeta^t=0$, where $\vx\to\mathcal{C}$ means that $\vx$ is classified in $\mathcal{C}$. However, as we will show later, choosing $\zeta^t=0$ is not guaranteed to be optimal, even though it seems rather intuitive. Similarly, as discussed in the introduction, the arbitrary choice $y=\pm 1$ is not necessarily the best. To visualize that, one is interested in understanding the asymptotic behavior of the score function $f$. Before stating the theorem giving these asymptotics, one must introduce two fundamental small-dimensional quantities that will be of particular interest.
\begin{itemize}
    \item The data matrix $\mcM=\mM^\trans\mM$, with $\mM = \left[\boldsymbol{\mu}_1^{1},\boldsymbol{\mu}_2^{1},\dots,\boldsymbol{\mu}_2^{T}\right]$. It is worth to note that even though $\mM$ is not accessible in practice, the data matrix $\mcM \in \mathbb{R}^{2T\times 2T}$ can be estimated consistently (see Section~\ref{app:estimation} of the appendix).
    \item The parameter matrix $\mcA$, which is the solution of the following system of equations:
    \begin{equation*}
        \left\{
            \begin{array}{lr}
                \forall t, ~ \delta^{t}=\frac{1}{T}\mcA_{tt},\\
                \mcA=\tilde{\mathbf{\Lambda}} + \tilde{\mathbf{\Lambda}}\left(\diag_{\bar{\boldsymbol{\delta}}}^{-1}-\tilde{\mathbf{\Lambda}}\right)^{-1}\tilde{\mathbf{\Lambda}}
            \end{array}
        \right.
    \end{equation*}
    with $\tilde{\delta}_j^t = \frac{\rho_j^t}{Tc(1-\delta^t)}$ and $\bar{\delta}^t = \tilde{\delta}_1^t + \tilde{\delta}_2^t$. \\
    The equations leading to $\mcA$ translates how the parameters and hyperparameters of each task interact with each other in the optimization problem.
\end{itemize}

These two quantities are mixed up in the following matrix:
\begin{equation*}
    \mTheta_0 = \left(\mcA\otimes \mathbb{1}_2\mathbb{1}_2^\trans\right) \odot \mcM
\end{equation*}

As long as uncertain labeling is not taken into account, it is natural to consider that labels are equals for datapoints in the same class, \ie $\forall \vx_i^t\in\mathcal{C}_j^t, y_i^t=\tilde{y}_j^t\in\mathbb{R}$. Therefore we define $\tilde{\vy}=[\tilde{y}_1^1,\tilde{y}_2^1,\tilde{y}_1^2,\dots,\tilde{y}_2^T]\in\mathbb{R}^{2T}$, which sums up the label values. Precisely, we have $\vy_\ell = \mD \tilde{\vy}$, with \begin{equation}
\label{eq:simple_degrees}
\mD = \sum\limits_{t=1}^T \mE_{tt}^{[T]}\otimes \begin{pmatrix}
  \mathbb{1}_{n_{\ell 1}^t} & \mathbb{0}_{n_{\ell 1}^t} \\
  \mathbb{0}_{n_{\ell 2}^t} & \mathbb{1}_{n_{\ell 2}^t}
 \end{pmatrix}
\end{equation}
In this specific case, the statistics of the score function are given by the following theorem:

\begin{theorem}
\label{th:main}
Under Assumptions \ref{ass:data_distribution_2} and \ref{ass:growth_rate}, and if labels are given by \eqref{eq:simple_degrees}, for any unlabeled sample $\vx\in\mathcal{C}_j^{t}$, and $f$ being its associated score,
\begin{equation*}
f\rightarrow \mathcal{N}\left(m_j^{t},{\sigma^{t}}^2\right)
\end{equation*}
with $m_j^t = (1-\delta^t){\va_j^t}^\trans\tilde{\vy}$ and $\sigma^t = (1-\delta^t)\sqrt{\tilde{\vy}^\trans\mB^t\tilde{\vy}}$
\begin{align*}
    \va_j^t&=\left({\ve_{t,j}^{[2T]}}^\trans\left(\mTheta-\frac{Tc\delta^t}{\rho^t}\mGamma\right) \diag_{\tilde{\vdelta}} \diag_{\veta}\right)^\trans \\
    \mB^t&=\diag_{\veta}\left[2\diag_{\tilde{\vdelta}}\left(\mTheta-\frac{Tc\delta^t}{\rho^t}\mGamma\right) - \mGamma^t\right] \diag_{\vr^t}\diag_{\veta} \\
    &+\diag_{\veta}\diag_{\tilde{\vdelta}}\left(\mTheta\diag_{\vr^t}\mTheta+\bar{\mOmega}^t-\left(\frac{Tc\delta^t}{\rho^t}\right)^2\mGamma^t\right)\diag_{\tilde{\vdelta}} \diag_{\veta} \\
    &+\mT^t\odot\diag_{\vrho \odot \veta}\diag_{(\bar{\vrho}\odot\bar{\veta})\otimes\mathbb{1}_2}^{-1}
\end{align*}
and where
\begin{itemize}
    \item $\mTheta = \left(\mI_{2T}-\mTheta_0\diag_{\tilde{\vdelta}}\right)^{-1}\mTheta_0 \in\mathbb{R}^{T\times T}$
    \item $\mGamma = \mI_T \otimes \mathbb{1}_2\mathbb{1}_2^\trans \in\mathbb{R}^{T\times T}$ and $\mGamma^t = \mE_{tt} \otimes \mathbb{1}_2\mathbb{1}_2^\trans \in\mathbb{R}^{T\times T}$
    \item $\bar{\mOmega}^t = \left(\mI_{2T}-\mTheta_0\diag_{\tilde{\vdelta}}\right)^{-1}\bar{\mOmega}_0^t\left(\mI_{2T}-\diag_{\tilde{\vdelta}}\mTheta_0\right)^{-1} \in\mathbb{R}^{2T\times 2T}$
    \item $\bar{\mOmega}_0^t = \left[\left(\mcA\diag_{\ve_t^{[T]}+\bar{\vr}^t}\mcA\right)\otimes \mathbb{1}_2\mathbb{1}_2^\trans\right] \odot \mcM \in\mathbb{R}^{2T\times 2T}$
    \item $\mT^t = \diag_{\bar{\vr}^t\odot\bar{\veta}}\otimes \mathbb{1}_2\mathbb{1}_2^\trans \in\mathbb{R}^{2T\times 2T}$
    \item $\vr^t=\vrho\odot\left(\mS_{t.}\otimes\mathbb{1}_2\right)\in\mathbb{R}^{2T}$ and $\bar{\vr}^t=\bar{\vrho}\odot\mS_{t.}\in\mathbb{R}^{T}$
    \item $\mS=\bar{\mS}\left(\mI_T-\diag_{\bar{\vrho}}\bar{\mS}\right)^{-1}\in\mathbb{R}^{T\times T}$ with $\bar{\mS}_{tt'}=\frac{\mcA_{tt'}^2}{T^2 c (1-\delta^{t'})^2}$
\end{itemize}
\end{theorem}

Beyond the cumbersome formulas, one can notice that the vector $\va_j^t$ and the matrix $\mB^t$ are small dimensional deterministic quantities that only depend on estimates, known parameters and hyperparameters. Interestingly, even without analysing how $\va_j^t$ and $\mB^t$ relate to the parameters of the problem, one can take a look at the implication of the Theorem \ref{th:main} on our classification problem. The key information here is that the decision score $f$ is asymptotically Gaussian with known parameters. These parameters are mere functionals of the label vector $\tilde{\vy}$. Therefore, a good choice of $\tilde{\vy}$ as well as $\zeta^t$ could lead to a lower classification error. First of all, we need to precise which quantity we want to minimize.

\begin{definition}
\label{def:error}
For a given target task $t$, the classification error of any unlabeled sample $\vx \in \mathcal{C}_1^t$ is
\begin{equation*}
    \epsilon_1^t=\P(\vx\to\mathcal{C}_2^{t}|\vx\in\mathcal{C}_1^{t})
\end{equation*}
Similarly, the classification error of any unlabeled sample $\vx \in \mathcal{C}_2^t$ is 
\begin{equation*}
    \epsilon_2^t=\P(\vx\to\mathcal{C}_1^{t}|\vx\in\mathcal{C}_2^{t})
\end{equation*}
\end{definition}

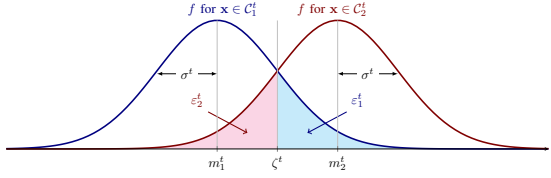
\begin{figure}[!t]
\centering
\begin{tikzpicture}[scale=0.5]
\begin{axis}[no markers, domain=-4.5:4.5, samples=100, hide y axis,
axis x line=bottom, height=5cm, width=16cm, xtick={-1,0,1}, xticklabels={$m_1^t$,$\zeta^t$,$m_2^t$}, ytick=\empty, enlargelimits=false, clip=false, axis on top, grid = major]
  \addplot[fill=cyan!20, draw=none, domain=0:3] {gauss(-1,1)} \closedcycle;
  \addplot[fill=magenta!20, draw=none, domain=-3:0] {gauss(1,1)} \closedcycle;
  \addplot[very thick,blue!50!black] {gauss(-1,1)} node [pos=0.4,anchor=south]
{$f$ for $\vx\in\mathcal{C}_1^t$};
  \addplot[very thick,red!50!black] {gauss(1,1)} node [pos=0.6,anchor=south]
{$f$ for $\vx\in\mathcal{C}_2^t$};
  \draw[red!50!black,<-,thick] (axis cs:-0.5,0.05) -- (axis cs:-1.15,0.115) node[anchor=south east]{$\epsilon_2^t$};
  \draw[blue!50!black,<-,thick] (axis cs:0.5,0.05) -- (axis cs:1.15,0.115) node[anchor=south west]{$\epsilon_1^t$};
  \draw [yshift=2cm,latex-latex](axis cs:-2,0) -- node [fill=white] {$\sigma^t$} (axis cs:-1,0);
  \draw [yshift=2cm,latex-latex](axis cs:1,0) -- node [fill=white] {$\sigma^t$} (axis cs:2,0);
\end{axis}
\end{tikzpicture}
\caption{Asymptotic probability distribution of the score function $f$ for samples of both classes. The classification errors expressed in Remark \ref{rm:error} can be interpreted as the area delimited by the density curve of $f$ and the threshold $\zeta^t$.}
\label{fig:error}
\end{figure}

\begin{remark}
\label{rm:error}
According to Theorem \ref{th:main}, and using the classification rule \eqref{eq:classif}, the classification errors of Definition \ref{def:error} rewrite as:
\begin{equation}
    \epsilon_1^t=\mathcal{Q}\left(\frac{\zeta^{t}-m_1^{t}}{\sigma^{t}}\right) \quad \text{and} \quad \epsilon_2^t=\mathcal{Q}\left(\frac{m_2^{t}-\zeta^{t}}{\sigma^{t}}\right)
\end{equation}
with $\mathcal{Q}(x)=\frac{1}{\sqrt{2\pi}}\int_{x}^{\infty}e^{-\frac{u^2}{2}} \,\mathrm{d}u$. Thus, $\zeta^t \mapsto \epsilon_1^t$ is a decreasing function, while $\zeta^t \mapsto \epsilon_2^t$ is an increasing function.
\end{remark}
With this remark, we understand that the choice of the threshold $\zeta^t$ is a trade-off between the minimization of $\epsilon_1^t$ and $\epsilon_2^t$, as displayed in Figure \ref{fig:error}.

 $\zeta^t=m_1^t$ and $\zeta^t=m_2^t$ are extreme choices, where we have respectively $\epsilon_1^t=\frac 12$ and  $\epsilon_2^t=\frac 12$. If $\zeta^t\notin [m_1^t,m_2^t]$, we have either $\epsilon_1^t>\frac 12$ or $\epsilon_2^t>\frac 12$, which is not a desirable situation. So we will assume starting from now that $\zeta^t \in [m_1^t,m_2^t]$. Interestingly, for any such choice of $\zeta^t$, the optimal label vector is the same.

\begin{proposition}
\label{th:optimal}
For a given target task $t$ and a given threshold $\zeta^t\in[m_1^t,m_2^t]$, there exists a unique (up to a multiplicative constant) score vector $\tilde{\vy}^\star$ minimizing $\epsilon_1^t$ and $\epsilon_2^t$, given as:
\begin{equation}
\label{eq:yopt}
    \tilde{\vy}^\star = (\mB^t)^{-1}(\va_2^t-\va_1^t)
\end{equation}
\end{proposition}

This property provides the possibility, depending on the use of the algorithm, to choose the threshold accordingly. Indeed, the choice of $\zeta^t$ can be made in several different ways, allowing the user to tackle a wide variety of classification problems. For example :
\begin{itemize}
    \item In a medical context, one could be interested to minimize the number of false positive $\epsilon_1^t$ while ensuring that the number of false negative $\epsilon_2^t$ is lower than a given threshold $p$ (for example $p=1\%$). Then the optimal $\zeta^t$ is given by: 
    $$\zeta^t = m_2^t - \sigma^t\mathcal{Q}^{-1}(p)$$ 
    \item One could also be interested in minimizing the overall classification error of the dataset $\frac{n_{u1}^t}{n_u^t}\epsilon_1^t+\frac{n_{u2}^t}{n_u^t}\epsilon_2^t$ (and therefore prioritize the classification of the most numerous class). Then the optimal $\zeta^t$ is given by: 
    $$\zeta^t = \frac{m_1^{t}+m_2^{t}}{2} + \frac{{\sigma^{t}}^2}{m_2^{t}-m_1^{t}}\log\left(\frac{n_{u1}^{t}}{n_{u2}^{t}}\right)$$
    \item If one wants to minimize undifferently $\epsilon_1^t$ and $\epsilon_2^t$ (so that $\epsilon_1^t=\epsilon_2^t$), then the optimal $\zeta^t$ is straightforwardly given by:
    $$\zeta^t = \frac{m_1^{t}+m_2^{t}}{2}$$
    This situation is a specific case of the previous one with $n_{u1}^t=n_{u2}^t$. It can also be interpreted as minimizing the overall classification error without any \textit{a priori} on the genuine class of unlabeled samples. We will focus on this last simple example to provide a lower bound of the classification error.
\end{itemize}

\begin{proposition}
\label{th:error}
For a given target task $t$, the minimal value of the classification error $\epsilon^t=\frac{\epsilon_1^t+\epsilon_2^t}{2}$ is achieved with the optimal label $\tilde{\vy}^{\star}$, and is asymptotically given by:
\begin{equation}
\label{eq:error}
    \epsilon_\star^t=\mathcal{Q}\left(\frac{1}{2}\sqrt{(\va_2^t-\va_1^t)^\trans (\mB^t)^{-1}(\va_2^t-\va_1^t)}\right),
\end{equation}
\end{proposition}
The proof of Propositions \ref{th:optimal} and \ref{th:error} are obtained from classical convex optimization tools and provided for simplicity of exposition in the Sections \ref{app:optimal} and \ref{app:error} of the appendix. We now see that the Theorem \ref{th:main} provides us:
\begin{itemize}
    \item An optimal choice $\tilde{\vy}^\star$ of the label vector $\tilde{\vy}$
    \item The information we need to choose wisely the threshold $\zeta^t$
    \item The optimal classification error $\epsilon_\star^t$ for a specific choice of $\zeta^t$ 
\end{itemize}

This leads to Algorithm \ref{alg:opt}, which summarizes the previous results and remarks.

\begin{algorithm}[h]
\caption{Optimal algorithm}
\label{alg:opt}
\begin{algorithmic}
    \STATE {{\bfseries Input:} labeled data $\mX_\ell = [\mX_\ell^{1},\dots,\mX_\ell^{T}]$ and unlabeled data $\mX_u = [\mX_u^{1},\dots,\mX_u^{T}]$}
    \STATE {{\bfseries Output:} Estimated class $\hat{j} \in \{1,2\}$ for unlabeled data of a given target task $t$}
    \STATE {\bfseries Center} data per task 
    \STATE {\bfseries Compute} data matrices $\mZ_\ell$ and $\mZ_u$ from $\mX_\ell$ and $\mX_u$
    \STATE {\bfseries Estimate} matrix $\mathcal M$ (cf. Section \ref{app:estimation})
    \STATE {\bfseries Create} scores $\tilde{\vy}^\star$ from \eqref{eq:yopt} and $\Lambda$ from \eqref{eq:lambda}.
    \STATE {\bfseries Estimate} the classification error $\epsilon_\star^t$ according to \eqref{eq:error} and optimize with respect to $\alpha$ using a grid search approach.
    \STATE {\bfseries Compute} classification scores $\vf_u$ according to \eqref{eq:solution}.
    \STATE {{\bfseries Output: }} $\hat j$ such that $f_i\underset{\hat j=1}{\overset{\hat j=2}{\gtrless}} \frac{m_1^{t}+m_2^{t}}{2}$.
\end{algorithmic}
\end{algorithm}

\section{Uncertain labeling}
\label{sec:uncertain}

In this section, we aim to extend the previous results to the case of erroneous or uncertain data. Until now, the value of the label $y_i^t$ of a given sample was either $\tilde{y}_1^t$ or $\tilde{y}_2^t$, depending wether the sample was belonging to the class $\mathcal{C}_1^t$ or $\mathcal{C}_2^t$. Let us consider now that instead of knowing which class the labeled sample belongs to, we have a couple $(d_{i1}^t, d_{i2}^t)$ of pre-estimated probabilities that the sample belongs to a class or to the other. The previous case is met if the couple is either $(1,0)$ or $(0,1)$, the associated label value being $\tilde{y}_1^t$ or $\tilde{y}_2^t$. For any other case, the value of $y_i^t$ should be in the interval $]\tilde{y}_1^t,\tilde{y}_2^t[$. A natural choice is to set $y_i^t = d_{i1}^t\tilde{y}_1^t + d_{i2}^t\tilde{y}_2^t$. Therefore, we have $\vy_\ell = \mD \tilde{\vy}$, with \begin{equation}
\label{eq:simple_degrees}
\mD = \sum\limits_{t=1}^T \mE_{tt}^{[T]}\otimes \begin{pmatrix}
    d_{11}^t & d_{12}^t \\
    d_{21}^t & d_{22}^t \\
    \vdots & \vdots \\
    d_{n_{\ell}^t 1} & d_{n_{\ell}^t 2}
 \end{pmatrix}
\end{equation}

The results of Theorem \ref{th:main} have to be adapted. In particular, we have to introduce some statistics related to the couples of probabilities :
\begin{itemize}
    \item $\bar{d}_{j_1,j_2}^t = \frac{1}{n_{\ell j_1}^t}\sum_{i'|x_{i'}\in \mathcal{C}_{j_1}^{t}} d_{i'j_2}^t$ the average probability for a genuine sample of class $\mathcal{C}_{j_1}^{t}$ to be labeled in class $\mathcal{C}_{j_1}^{t}$. In the ideal case, $\bar{d}_{j_1,j_2}^t = \mathbb{1}_{j_1=j_2}$.
    \item $\tilde{d}_{j_1j_2}^t = \frac{1}{n_\ell^t}\sum_{i|\vx_{i}\in\mathcal{C}^{t}} d_{ij_1}^t d_{ij_2}^t$ which is a measure of the labeling uncertainty. If $j_1\neq j_2$ (resp. $j_1=j_2$), a high value of $\tilde{d}_{j_1j_2}^t$ means a high (resp. low) uncertainty.
\end{itemize}

This leads to the following small-dimensional quantities, related to the matrix $\mD$ :
\begin{align*}
    \bar{\mD}&=\sum_{t=1}^T \mE_{tt}\otimes \begin{pmatrix}
            \bar{d}_{11}^t & \bar{d}_{12}^t\\
            \bar{d}_{21}^t & \bar{d}_{22}^t \end{pmatrix}, \\
    \tilde{\mD}&=\sum_{t=1}^T \mE_{tt}\otimes \begin{pmatrix}
            \tilde{d}_{11}^t & \tilde{d}_{12}^t\\
            \tilde{d}_{21}^t & \tilde{d}_{22}^t \end{pmatrix}.
\end{align*}

\begin{theorem}
\label{th:main_bis}
Under Assumptions \ref{ass:data_distribution_2} and \ref{ass:growth_rate}, and if labels are given by \eqref{eq:simple_degrees}, for any unlabeled sample $\vx\in\mathcal{C}_j^{t}$, and $f$ being its associated score,
\begin{equation*}
f\rightarrow \mathcal{N}\left(m_j^{t},{\sigma^{t}}^2\right)
\end{equation*}
with $m_j^t = (1-\delta^t){\va_j^t}^\trans\tilde{\vy}$ and $\sigma^t = (1-\delta^t)\sqrt{\tilde{\vy}^\trans\mB^t\tilde{\vy}}$
\begin{align*}
    \va_j^t&=\left({\ve_{t,j}^{[2T]}}^\trans\left(\mTheta-\frac{Tc\delta^t}{\rho^t}\mGamma\right) \diag_{\tilde{\vdelta}} \diag_{\veta}\bar{\mD}\right)^\trans \\
    \mB^t&=\bar{\mD}^\trans\diag_{\veta}\left[2\diag_{\tilde{\vdelta}}\left(\mTheta-\frac{Tc\delta^t}{\rho^t}\mGamma\right) - \mGamma^t\right] \diag_{\vr^t}\diag_{\veta}\bar{\mD} \\
    &+\bar{\mD}^\trans\diag_{\veta}\diag_{\tilde{\vdelta}}\left(\mTheta\diag_{\vr^t}\mTheta+\bar{\mOmega}^t-\left(\frac{Tc\delta^t}{\rho^t}\right)^2\mGamma^t\right)\diag_{\tilde{\vdelta}} \diag_{\veta}\bar{\mD} \\
    &+\mT^t\odot\tilde{\mD}
\end{align*}
\end{theorem}

The Theorem \ref{th:main} is a particular case of the Theorem \ref{th:main_bis}, with $\bar{\mD}=\mI_{2T}$ and $\tilde{\mD}=\diag_{\vrho \odot \veta}\diag_{(\bar{\vrho}\odot\bar{\veta})\otimes\mathbb{1}_2}^{-1}$.

\section{Limitations}
\label{sec:limitations}

\subsection{Multiclass setting}
\label{sec:multiclass}
The paper focuses on a binary setting. There are some known methods to deal with a higher number of classes, such as \emph{one-vs-all} and \emph{one-vs-one} \cite{bishop,rocha_2014}. These methods can be roughly implemented in our case, but each one has some limitations. In this paragraph, $m>2$ will denote the number of classes.
\begin{itemize}
    \item \textbf{One-vs-all} : $m$ binary classifiers are trained, each one separating one class $\mathcal{C}_j$ from the other $m-1$ classes. Each new sample is then classified into the class with the highest score among the $m$ classifiers. The computations of Theorem \ref{th:main} can be easily adapted to these classifiers, but this approach has several issues :
        \begin{itemize}
            \item Each classifier has its own bias, making impossible a fair comparison between the output scores. To work well, this method implicitly makes the assumption that for all $1 \leq j \leq m$, the output distribution of the classifier $\mathcal{C}_j$-vs-all on samples from class $\mathcal{C}_j$ is the same. One way to tackle this problem is to normalize the distribution (\ie consider $\frac{f_i^t-m_j^t}{\sigma^t}$ instead of $f_i^t$), but the algorithm then depends too heavily on the estimations of $m_j^t$ and $\sigma^t$, affecting its robustness.
            \item The optimal score vector $\tilde{\vy}^\star$ does not have an explicit formula, as the quantity to maximize is not a quadratic form of $\tilde{\vy}$ anymore (see Section \ref{app:optimal} of the appendix). As such, one needs to approximate the quantity to maximize, therefore losing the optimality.
        \end{itemize}
    \item \textbf{One-vs-one} : $\frac{1}{2}m(m-1)$ binary classifiers are trained, each one separating one class $\mathcal{C}_j$ from an other class $\mathcal{C}_{j'}$. For every sample of unlabeled data, each one of these classifiers "votes" for the class the most relevant among the two that it compares. Then, the sample is attributed to the class collecting the most votes. This approach also comes with several issues :
        \begin{itemize}
                \item To train a binary classifier $\mathcal{C}_j$-vs-$\mathcal{C}_{j'}$, one can get rid of labeled data from other classes, but not unlabeled data. As such, the unsupervised part of the algorithm will still make use of the unlabeled data of other classes and change the outcome of the score function, necessarily introducing a new bias. However, results of Theorem \ref{th:main} can be adapted to this situation.
                \item The number of one-vs-one classifiers to train is a quadratic function of the number of classes, while the number of one-vs-all classifiers was a linear function of the number of classes. With a high number of classes, the one-vs-one method is therefore much slower.
        \end{itemize}
\end{itemize}
To our knowledge, one-vs-one is the only method able to keep the optimality we have in the binary case. The major issue is that a high number of classes makes the method heavy to use.

\subsection{Lack of labeled data}

Even if the model is designed to work for any number of labeled and unlabeled data, in practice it cannot do without a minimum number of labeled data. First of all, if there is no labeled data at all (as it would be the case in a purely unsupervised setting), the score vector $\vf$ would be zero as the label vector $\vy$ would be zero itself.
Most importantly, labeled data is needed to perform the estimation of $\mcM$ and $\mLambda$. If labeled data is too scarce, the algorithm could be misleaded by the estimated values of $\mcM$ and $\mLambda$, implying poor performances.

\section{Experiments}
\label{sec:exp}

The experiments displayed in this section are done on both synthetic and real-world datasets. With Gaussian synthetic data, we simulate different learning settings, in order to convey some intuition on the different aspects of our model : multi-task learning, semi-supervised learning, uncertain labeling, floating labels... Meanwhile, the purpose of real-data experiments is to show that our algorithm is robust enough to keep its properties on dataset that are not isotropic Gaussian vectors anymore. The code used to genereate the following experiments is available at \href{https://gricad-gitlab.univ-grenoble-alpes.fr/legervi/tsp}{https://gricad-gitlab.univ-grenoble-alpes.fr/legervi/tsp}

\subsection{Multitask experiments}
\label{sec:multi_task}

This section investigates the influence of task correlation on the performances of our algorithm. To do so, we consider a case of transfer learning between a source task and a target task. The source and target tasks are mixtures of two Gaussians: $\mathcal{N}(\pm\vmu,\mI_p)$ for the source and $\mathcal{N}(\pm(\beta\vmu+\sqrt{1-\beta^2}\vmu^{\perp}),\mI_p)$ for the target, where $\vmu^\perp$ is a vector orthogonal to the vector $\vmu$. This setting allows through $\beta$ to control the similarity between both tasks. Specifically, for $\beta=0$ the tasks are unrelated while for $\beta=1$ they are identical. Note that $\beta=-1$ corresponds to a scenario where the classes of target and source tasks are reversed.

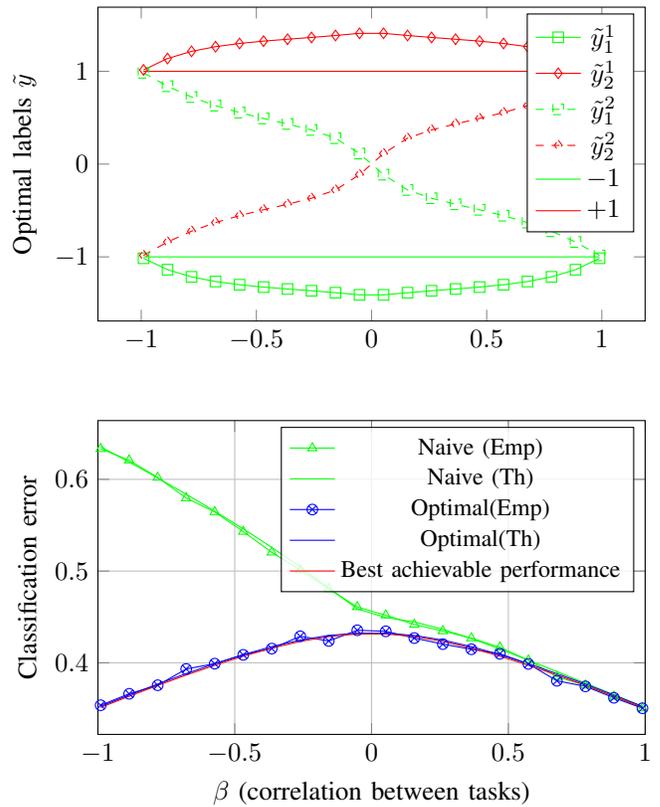
\begin{figure}[!t]
\centering
\begin{tikzpicture}
\begin{axis}[grid=major,ylabel={Classification error},xmin=-1,xmax=1,xlabel={$\beta$ (correlation between tasks)},width=1\linewidth,height=0.65\linewidth,legend={columns=1},legend style={fill opacity=0.8,text opacity=1,font=\small}]
    \addplot[thin,mark=triangle,color=green]table[x=beta,y=m_naive_emp] {correlation_perf_2.dat};
    \addplot[thin,color=green]table[x=beta,y=m_naive_th] {correlation_perf_2.dat};
    \addplot[thin,mark=otimes,color=blue]table[x=beta,y=m_opt_emp] {correlation_perf_2.dat};
    \addplot[thin,color=blue]table[x=beta,y=m_opt_th] {correlation_perf_2.dat};
    \addplot[red] table[x=beta,y=best] {correlation_perf_2.dat};
    \legend{Naive (Emp),Naive (Th),Optimal(Emp),Optimal(Th),Best achievable performance}
\end{axis}
\begin{axis}[yshift=5.5cm,ylabel={Optimal labels $\tilde{y}$},width=1\linewidth,height=0.65\linewidth]
    \addplot[green, mark=square] table[x=beta,y=y_opt_1_1] {correlation_labels_2.dat};
    \addlegendentry{$\tilde{y}_1^{1}$}
    \addplot[red, mark=diamond] table[x=beta,y=y_opt_1_2] {correlation_labels_2.dat};
    \addlegendentry{$\tilde{y}_2^{1}$}
    \addplot[green, mark=square, dashed] table[x=beta,y=y_opt_2_1] {correlation_labels_2.dat};
    \addlegendentry{$\tilde{y}_1^{2}$}
    \addplot[red, mark=diamond, dashed] table[x=beta,y=y_opt_2_2] {correlation_labels_2.dat};
    \addlegendentry{$\tilde{y}_2^{2}$}
    \addplot[green] table[x=beta,y=y_naive_1] {correlation_labels_2.dat};
    \addlegendentry{$-1$}
    \addplot[red] table[x=beta,y=y_naive_2] {correlation_labels_2.dat};
    \addlegendentry{$+1$}
\end{axis}
\end{tikzpicture}
\caption{Joint evolution of optimal labeling and classification error as a function of correlation between tasks ($p=200$, $n_\ell^1=100$, $n_\ell^2=1000$, $n_u^1=n_u^2=250$). (\textbf{Top}) Optimal labels with normalization $\|\tilde{\vy}\|=1$. Optimal labels adapt themselves to avoid negative transfer (\textbf{Bottom}) Classification error for both naive and optimal algorithms. Our algorithm is close to optimal, while naive labels induce a negative transfer when tasks are not related enough.}
\label{fig:correlation}
\end{figure}

Figure \ref{fig:correlation} displays the empirical versus theoretical classification error of our algorithm as a function of $\beta$ for the optimized method (which integrates the label and hyperparameter optimization) against the naive method which leaves the labels ($\pm1$) untouched. The values of the optimal labels are displayed jointly in the top figure.

We also provide an optimal bound of performance established by information theory in \cite{nguyen2023asymptotic}. This bound corresponds to the best performance achievable by any algorithm given the setting of our experiment. The figure clearly shows that our proposed optimized algorithm is extremely close to the information-theoretic optimum under our Assumptions \ref{ass:data_distribution_2} and \ref{ass:growth_rate}. The close fit between theoretical and empirical performances confirms Proposition~\ref{th:error} even for finite-dimensional data.

We also observe a strong robustness of our method to negative transfer. Indeed, when $\beta$ becomes negative, the error increases for the naive algorithm, while our proposed method is insensitive to the sign of $\beta$ by a dynamic adaptation of the labels. The top figure shows the sign inversion of $\tilde{y}_1^2$ and $\tilde{y}_2^2$ when $\beta$ becomes negative. In the specific case where $\beta=-1$, our algorithm has the same performance as with $\beta=1$, while the naive ``conventional'' approach gets worse than random guess. Besides, when $\beta$ is close to $0$, the labels of the source task have low magnitude, because there is no much information to get from source task.

\subsection{Class imbalances}

In the previous experiment, the number of data in each class was the same. However, our adaptive labeling is also useful to deal with class imbalances. Indeed, labels can be interpreted as weights, which should naturally be higher for underrepresented classes.
We will consider a single-task learning scenario with a fixed amount of data. While the number of unlabeled data is equal for both classes, the number of labeled data is unbalanced.

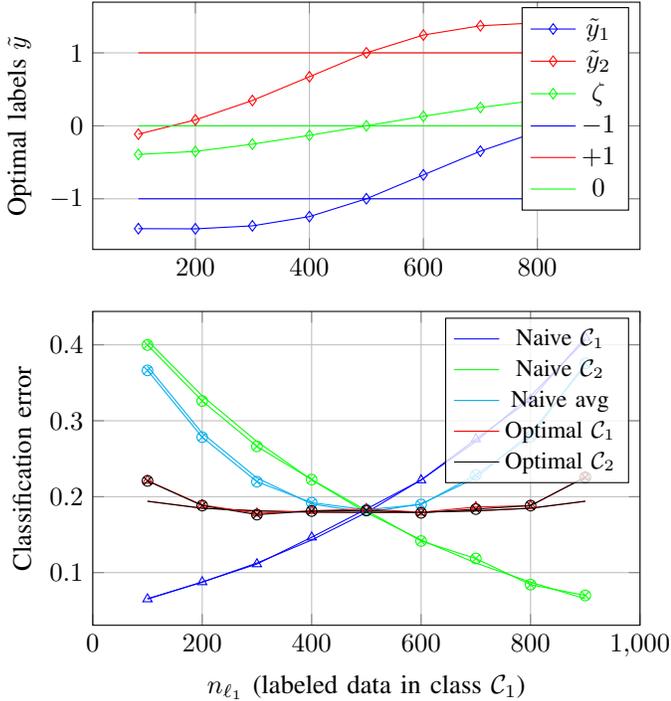
\begin{figure}[!t]
\centering
\begin{tikzpicture}
\begin{axis}[grid=major,yshift=5cm,ylabel={Optimal labels $\tilde{y}$},width=1\linewidth,height=0.55\linewidth],legend style={fill opacity=0.8,text opacity=1,font=\small}]
    \addplot[blue, mark=diamond] table[x=nl,y=y_opt_1] {imbalances_labels.dat};
    \addlegendentry{$\tilde{y}_1$}
    \addplot[red, mark=diamond] table[x=nl,y=y_opt_2] {imbalances_labels.dat};
    \addlegendentry{$\tilde{y}_2$}
    \addplot[green, mark=diamond] table[x=nl,y=threshold_opt] {imbalances_labels.dat};
    \addlegendentry{$\zeta$}
    \addplot[blue] table[x=nl,y=y_naive_1] {imbalances_labels.dat};
    \addlegendentry{$-1$}
    \addplot[red] table[x=nl,y=y_naive_2] {imbalances_labels.dat};
    \addlegendentry{$+1$}
    \addplot[green] table[x=nl,y=threshold_naive] {imbalances_labels.dat};
    \addlegendentry{$0$}
\end{axis}
\begin{axis}[grid=major,ylabel={Classification error},xmin=0,xmax=1000,xlabel={$n_{\ell_1}$ (labeled data in class $\mathcal{C}_1$)},width=1\linewidth,height=0.65\linewidth,legend={columns=1},legend style={fill opacity=0.8,text opacity=1,font=\small}]
    \addplot[thin,color=blue]table[x=nl,y=m_naive_th_1] {class_imbalances.dat};
    \addplot[thin,color=green]table[x=nl,y=m_naive_th_2] {class_imbalances.dat};
    \addplot[thin,color=cyan]table[x=nl,y=m_naive_th_avg] {class_imbalances.dat};
    \addplot[thin,color=red]table[x=nl,y=m_opt_th_1] {class_imbalances.dat};
    \addplot[thin,color=black]table[x=nl,y=m_opt_th_2] {class_imbalances.dat};
    
    \addplot[thin,mark=triangle,color=blue]table[x=nl,y=m_naive_emp_1] {class_imbalances.dat};
    \addplot[thin,mark=otimes,color=green]table[x=nl,y=m_naive_emp_2] {class_imbalances.dat};
    \addplot[thin,mark=otimes,color=cyan]table[x=nl,y=m_naive_emp_avg] {class_imbalances.dat};
    \addplot[thin,mark=triangle,color=red]table[x=nl,y=m_opt_emp_1] {class_imbalances.dat};
    \addplot[thin,mark=otimes,color=black]table[x=nl,y=m_opt_emp_2] {class_imbalances.dat};
    \legend{Naive $\mathcal{C}_1$, Naive $\mathcal{C}_2$, Naive avg, Optimal $\mathcal{C}_1$ , Optimal $\mathcal{C}_2$}
\end{axis}
\end{tikzpicture}
\caption{Joint evolution of optimal labeling and classification error as a function of the number of labeled data in class $\mathcal{C}_1$. The total number of labeled data is constant ($n_\ell=n_{\ell 1}+n_{\ell 2}=1000$, $p=200$, $n_{u1}=n_{u2}=200$). (\textbf{Top}) Optimal labels with normalization $\|\tilde{\vy}\|=1$, and optimal threshold $\zeta$ (also normalized). Optimal labels adapt themselves to compensate the class imbalances (\textbf{Bottom}) Theoretical and empirical classification error for both naive and optimal labels and threshold. The overall error is better with our algorithm, while naive labels and threshold induce a high error for the most represented class.}
\label{fig:imbalances}
\end{figure}

As such, Figure \ref{fig:imbalances} displays the classification error as a function of the number of labeled data in class $\mathcal{C}_1$, jointly with the values of the optimal labels (top figure). We also display the value of the optimal threshold $\zeta=\frac{m_1+m_2}{2}$, which ensures that the probability of misclassifying each sample is the same whether this sample is in a class or in the other. As expected, the label value of the least represented class is higher, as each labeled sample belonging to this class carries more information. Because of the imbalance, the means of the two classes are not symetrical about zero anymore, and the threshold value must be changed.

In the naive setting, without our choice of threshold, the classification error for samples from the most represented class is much higher. As a consequence, the average classification error is higher overall.

\subsection{Uncertain labeling}

\begin{figure}[!t]
\centering
\begin{tikzpicture}
\begin{axis}[grid=major,ylabel={$n_i$ (imprecise data)},xlabel={$n_r$ (reliable data)},width=1\linewidth,height=0.65\linewidth,legend={columns=1},legend style={fill opacity=0.8,text opacity=1,font=\small,legend pos=north west}]
    \addplot[thin,mark=diamond,color=blue]table[x=nl1,y=ni1] {nb_donnees_sup.dat};
    \addplot[thin,mark=diamond,color=red]table[x=nl2,y=ni2] {nb_donnees_sup.dat};
    \addplot[thin,mark=diamond,color=green]table[x=nl3,y=ni3] {nb_donnees_sup.dat};
    \addplot[thin,mark=diamond,color=magenta]table[x=nl4,y=ni4] {nb_donnees_sup.dat};
    \addplot[thin,mark=diamond,color=cyan]table[x=nl5,y=ni5] {nb_donnees_sup.dat};
    \legend{$r=0.5$, $r=0.6$, $r=0.75$, $r=0.9$, $r=1$}
\end{axis}
\end{tikzpicture}
\caption{Number of imprecise data $n_i$ needed to reach the same performance one had using $n_r$ samples of reliable data, for different values of $r$ ($1$-task, $p=200$, $n_{u1}=n_{u2}=200$). For each point, $n_r$ is a random number between $20$ and $400$. The figure strongly suggests that $n_i$ is a linear function of $n_r$.}
\label{fig:nb_donnees_sup}
\end{figure}
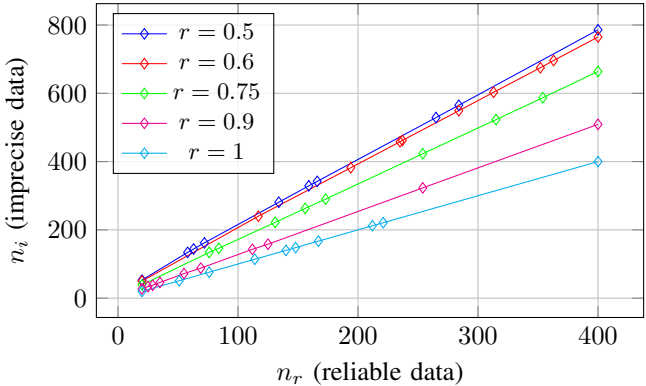

As for now, the labeled data used in the experiments was assumed to be labeled without any mistake or imprecision. To simulate a case where labeled data suffers such imprecision, we will separate the labeled data in two categories :
\begin{itemize}
    \item \textit{reliable} data, labeled in a class with absolute certainty.
    \item \textit{imprecise} data, for which there is a probability $r<1$ that the data genuinely belongs to the class it has been labeled in (and therefore a probability $1-r$ that it belongs in fact to the other class).
\end{itemize}
To simplify the setting, we assume that all the imprecise data has the same value $r$ of reliability. $n_i$ will denote the number of imprecise data, while $n_r$ will denote the number of reliable data. In order to measure how useful imprecise data is to solve our problem, one could ask the following question : for a given value of $n_r$, if imprecise data is used  instead of reliable data, how many samples $n_i$ are needed to reach the same performance ? The first thing we notice, through Figure \ref{fig:nb_donnees_sup}, is that the number of imprecise data needed to reach that performance is a linear function of the number of reliable data used in the first place.

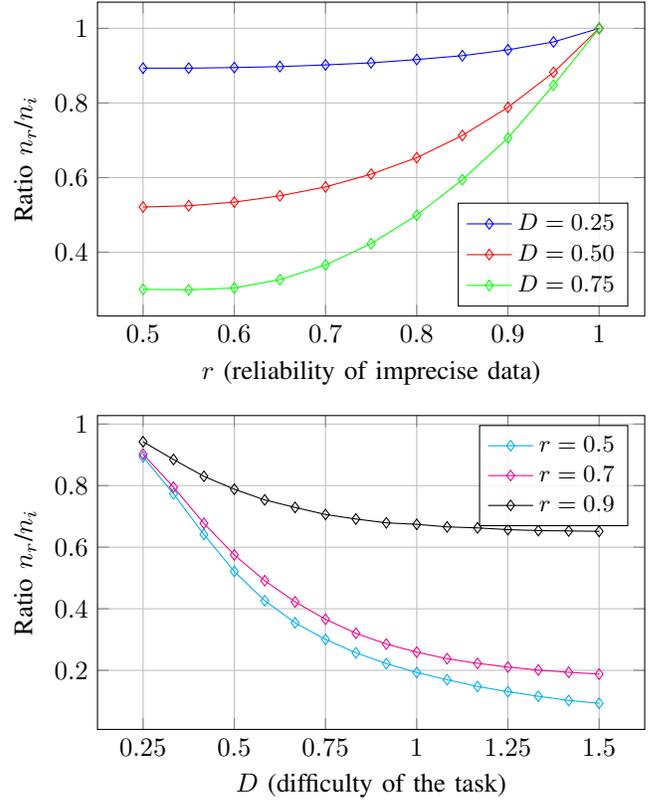
\begin{figure}[!t]
\centering
\begin{tikzpicture}
\begin{axis}[yshift=5.5cm,grid=major,ylabel={Ratio $n_r$/$n_i$},xlabel={$r$ (reliability of imprecise data)},width=1\linewidth,height=0.65\linewidth,legend={columns=1},legend style={fill opacity=0.8,text opacity=1,font=\small,legend pos=south east}]
    \addplot[thin,mark=diamond,color=blue]table[x=conf,y=ratio_1] {ratio_conf.dat};
    \addplot[thin,mark=diamond,color=red]table[x=conf,y=ratio_2] {ratio_conf.dat};
    \addplot[thin,mark=diamond,color=green]table[x=conf,y=ratio_3] {ratio_conf.dat};
    \legend{$D=0.25$, $D=0.50$, $D=0.75$}
\end{axis}
\begin{axis}[xtick={0.25,0.5,0.75,1,1.25,1.5},ytick={0,0.2,0.4,0.6,0.8,1},grid=major,ylabel={Ratio $n_r$/$n_i$},xlabel={$D$ (difficulty of the task)},width=1\linewidth,height=0.65\linewidth,legend={columns=1},legend style={fill opacity=0.8,text opacity=1,font=\small,legend pos=north east}]
    \addplot[thin,mark=diamond,color=cyan]table[x=diff,y=ratio_1] {ratio_diff.dat};
    \addplot[thin,mark=diamond,color=magenta]table[x=diff,y=ratio_2] {ratio_diff.dat};
    \addplot[thin,mark=diamond,color=black]table[x=diff,y=ratio_3] {ratio_diff.dat};
    \legend{$r=0.5$, $r=0.7$, $r=0.9$}
\end{axis}
\end{tikzpicture}
\caption{Ratio $\frac{n_r}{n_i}$ for different values of reliability ($r$) and difficulty of the task ($D=1/\|\vmu_1-\vmu_2\|$). The higher the ratio is, the more effective is the contribution of imprecise samples to the task. Both figures show that the harder the task is, the least useful imprecise samples are. However, an increasing of $r$ leads to significantely better results. (\textbf{Top}) Ratio $\frac{n_r}{n_i}$ as a function of the reliability of imprecise data, for different values of difficulty. (\textbf{Bottom}) Ratio $\frac{n_r}{n_i}$ as a function of the difficulty of the task, for different values of reliability.}
\label{fig:ratio}
\end{figure}

Therefore, the ratio $\frac{n_r}{n_i}$ seems to be a relevant quantity to compute. It can be interpreted as a measure of the strength of imprecise data compared to reliable data. The higher this ratio is, the more information is brought by imprecise samples. $\frac{n_r}{n_i}=0$ means that imprecise samples are not useful at all, while $\frac{n_r}{n_i}=1$ means that imprecise samples are as useful as reliable samples. Figure \ref{fig:ratio} displays this ratio for different values of reliability $r$ and difficulty of the task (expressed through the quantity $D=\frac{1}{\|\vmu_1-\vmu_2\|}$).

\subsection{Real data experiments}
\label{sec:real_data}

The purpose of this section is to check that our algorithm still performs well on real data, which does not necessarily follow Assumption \ref{ass:data_distribution_2}. The data is generated with normalized VGG-features \cite{simonyan2015deep} of randomly BigGAN-generated images \cite{brock2019large}. If GANs are known to produce concentrated vectors \cite{Seddik2020RandomMT}, the covariance is unlikely to be isotropic. However, thanks to the robustness of our method, we still have some reasonably good performances. In the following figures, only the empirical classification error is displayed, as the theoretical values are not relevant for this real data setting.

\begin{figure}[!t]
    \centering
	\begin{tabular}{cccc}
		\includegraphics[width=0.2\linewidth]{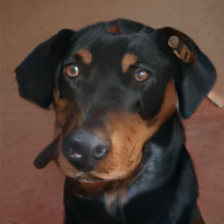} & \includegraphics[width=0.2\linewidth]{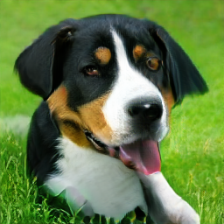} & \includegraphics[width=0.2\linewidth]{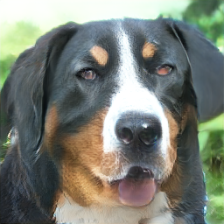} & \includegraphics[width=0.2\linewidth]{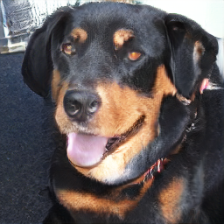} \\ \includegraphics[width=0.2\linewidth]{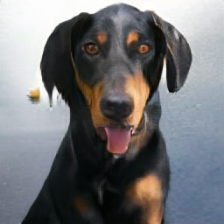} & \includegraphics[width=0.2\linewidth]{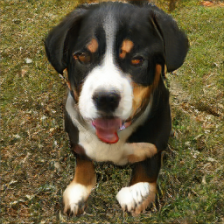} & \includegraphics[width=0.2\linewidth]{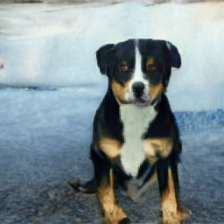} & \includegraphics[width=0.2\linewidth]{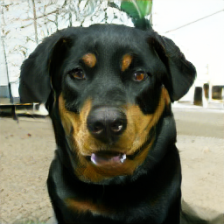} \\ \includegraphics[width=0.2\linewidth]{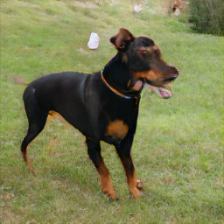} & \includegraphics[width=0.2\linewidth]{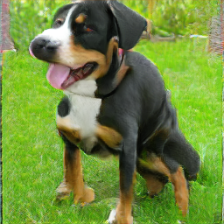} & \includegraphics[width=0.2\linewidth]{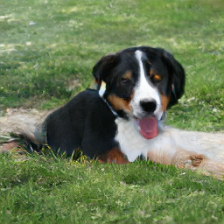} & \includegraphics[width=0.2\linewidth]{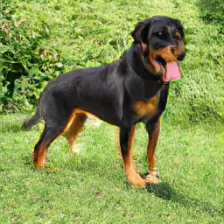}
	\end{tabular}
\caption{Examples of BigGAN-generated images used for the experiment. From left to right : \textit{doberman} (Class $\mathcal{C}_1^1$), \textit{entlebucher} (Class $\mathcal{C}_2^1$), \textit{appenzeler} (Class $\mathcal{C}_1^2$) and \textit{rottweiler} (Class $\mathcal{C}_2^2$). \textit{entlebucher} and \textit{appenzeler} are close, as well as \textit{doberman} and \textit{rottweiler}.}
\label{fig:images}
\end{figure}

\begin{figure}[!t]
\centering
\begin{tikzpicture}
\begin{axis}[grid=major,ylabel={Classification error},xlabel={$n_{\ell}^2$},width=1\linewidth,height=0.65\linewidth,legend={columns=1},legend style={fill opacity=0.8,text opacity=1,font=\small}]
    \addplot[color=green,error bars/.cd, x dir=both, x explicit, y dir=both, y explicit]table[x=nl,y=m_naive_emp,y error=s_naive_emp] {real_naive.dat};
    \addplot[color=blue,error bars/.cd, x dir=both, x explicit, y dir=both, y explicit]table[x=nl,y=m_opt_emp,y error=s_opt_emp] {real_naive.dat};
    \legend{Naive (Emp), Optimal (Emp)}
\end{axis}
\end{tikzpicture}
\caption{Empirical classification error, for both naive and optimal labels, as a function of the number of labeled data in source task, for BigGAN-generated images. (\textit{doberman} vs \textit{entlebucher} and \textit{appenzeler} vs \textit{rottweiler}, $p=4096$, $n_{\ell}^1=20$, $n_u^1=200$ and $n_u^2=0$). The naive labels induce a huge negative transfer, while optimal labels keep the error at low level.}
\label{fig:real_naive}
\end{figure}
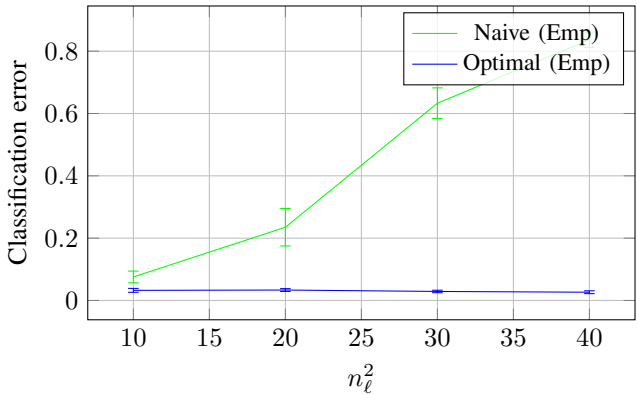

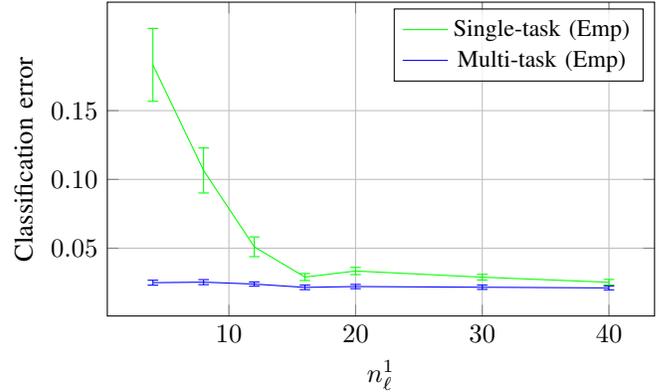
\begin{figure}[!t]
\centering
\begin{tikzpicture}
\begin{axis}[grid=major,ytick={0.05,0.10,0.15},yticklabels={$0.05$,$0.10$,$0.15$},ylabel={Classification error},xlabel={$n_{\ell}^1$},width=1\linewidth,height=0.65\linewidth,legend={columns=1},legend style={fill opacity=0.8,text opacity=1,font=\small}]
    \addplot[color=green,error bars/.cd, x dir=both, x explicit, y dir=both, y explicit]table[x=nl,y=m_single_emp,y error=s_single_emp] {real_multi_task.dat};
    \addplot[color=blue,error bars/.cd, x dir=both, x explicit, y dir=both, y explicit]table[x=nl,y=m_multi_emp,y error=s_multi_emp] {real_multi_task.dat};
    \legend{Single-task (Emp), Multi-task (Emp)}
\end{axis}
\end{tikzpicture}
\caption{Empirical classification error, for $1$-task and $2$-task learning, as a function of the number of labeled data in target task, for BigGAN-generated images. (\textit{doberman} vs \textit{entlebucher} and \textit{appenzeler} vs \textit{rottweiler}, $p=4096$, $n_{\ell}^2=100$, $n_u^1=200$ and $n_u^2=0$). The multi-task setting takes advantage of the labeled data from source task.}
\label{fig:real_multi_task}
\end{figure}

In the Figure \ref{fig:real_naive}, we consider a transfer-learning setting with $2$ similar tasks (\verb+doberman+ vs \verb+entlebucher+ and \verb+appenzeler+ vs \verb+rottweiler+, for which examples are displayed in Figure \ref{fig:images}) and with classes purposely inverted between the tasks (similarly to the case $\beta=-1$ of Section \ref{sec:multi_task}). Naturally, the naive algorithm suffers from a huge negative transfer, while optimal labeling ensures a low error. The same tasks are used in Figure \ref{fig:real_multi_task} to compare $1$-task and $2$-task learning (with our optimal algorithm). While the number of labeled data in the source task is constant, we add labeled data in the target task. The error of the single-task algorithm decreases, but remains higher than the multi-task algorithm, which benefits from the numerous data of the second task. To simplify the interpretation, the fully-supervised algorithm is considered.

\begin{figure}[!t]
\centering
\begin{tikzpicture}
\begin{axis}[grid=major,ytick={0.015,0.020,0.025},ylabel={Classification error},xlabel={$n_u$},width=1\linewidth,height=0.65\linewidth,legend={columns=1},legend style={fill opacity=0.8,text opacity=1,font=\small}]
    \addplot[color=green,error bars/.cd, x dir=both, x explicit, y dir=both, y explicit]table[x=nu,y=m_sup_emp,y error=s_sup_emp] {real_semi_sup.dat};
    \addplot[color=blue,error bars/.cd, x dir=both, x explicit, y dir=both, y explicit]table[x=nu,y=m_opt_emp,y error=s_opt_emp] {real_semi_sup.dat};
    \legend{Supervised (Emp), Semi-supervised (Emp)}
\end{axis}
\end{tikzpicture}
\caption{Empirical classification error for fully-supervised and semi-supervised algorithms, as a function of the number of unlabeled data. (BigGAN-generated images \textit{boxer} vs \textit{great swiss mountain dog}, $p=4096$, $n_{\ell}=10$). As expected, the error is constant with the fully-supervised algorithm, while the semi-supervised version benefits from additional unlabeled data.}
\label{fig:real_semi_sup}
\end{figure}
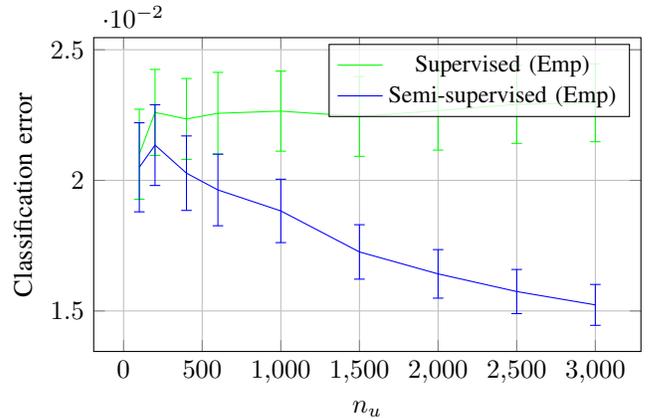

A third experiment is made with a semi-supervised setting, to which we compare the purely supervised method. Figure \ref{fig:real_semi_sup} shows the performances for the task \verb+boxer+ vs \verb+greater swiss mountain dog+ as a function of the number of unlabeled data. The error of the supervised method is logically constant, while the semi-supervised algorithm benefits from the new unlabeled samples.

\section{Concluding remarks}

Thanks to its simplicity, the algorithm presented in this article allows an extensive mathematical analysis, which makes it at once robust, powerful and interpretable. Indeed, the comparison of the empirical error with the optimum-theoretic bound indicates that the algorithm reaches satisfying performances, and the basic assumptions made still allow it to perform well on real data. 

Our algorithm is therefore accessible to its users and is even able to bring them further knowledge. More precisely :
\begin{itemize}
    \item the label vector $\tilde{\vy}$ informs the users on potential biases in the dataset, and helps them quantify how useful an additional task would be to solve the initial problem;
    \item the combined knowledge of the performance and the information-theoretic bound helps understand how difficult a problem is, and how far from optimal the associated performance lies;
    \item the experiments enlighten the users about the expected improvements of performance with additional labeled or unlabeled data, or when controversial data is used.
\end{itemize}
As such, our method does not intend to compete with state-of-the-art algorithms, but rather guarantees {\it sufficient} performances. In accordance with the base ``convivality'' principles enunciated in the introduction, our future investigations will not aim at any further complexification and improvement of the present method; rather, we will seek for further simplification and accessibility: the random matrix framework remains indeed quite opaque to most, which still constitutes an accessibility hurdle we should seek to overtake. Options lie in exploiting alternative tools from the mathematical, and even preferrably from the statistical physics, litterature in order to retrieve our technical results from a more direct and more intuitive path (which for instance the non-rigorous but much simpler replica methods may offer).

\appendices

\section{Solution of the optimization problem}
\label{app:optimization}

First of all, we need to clarify why \eqref{eq:opti_pb_2} is equivalent to \eqref{eq:opti_pb_3}.
\begin{align*}
    &\sum\limits_{t,t'=1}^T \Lambda^{tt'}\sum\limits_{i=1}^{n^{t}}\sum\limits_{i'=1}^{n^{t'}}\hat{\omega}_{ii'}^{tt'}\left(f_i^{t}-f_{i'}^{t'}\right)^2 \\
    =&\sum\limits_{t,t'=1}^T \Lambda^{tt'}\sum\limits_{i=1}^{n^{t}}\sum\limits_{i'=1}^{n^{t'}}\left[\hat{\omega}_{ii'}^{tt'}\left((f_i^{t})^2-(f_{i'}^{t'})^2\right)-2f_i^{t}\hat{\omega}_{ii'}^{tt'}f_{i'}^{t'}\right] \\
    =&\sum\limits_{t,t'=1}^T \Lambda^{tt'}\left(\sum\limits_{i=1}^{n^{t}}(f_i^{t})^2\underbrace{\sum\limits_{i'=1}^{n^{t'}}\hat{\omega}_{ii'}^{tt'}}_{=0}+\sum\limits_{i'=1}^{n^{t'}}(f_{i'}^{t'})^2\underbrace{\sum\limits_{i=1}^{n^{t}}\hat{\omega}_{ii'}^{tt'}}_{=0}\right) \\
    &-2\sum\limits_{t,t'=1}^T \Lambda^{tt'}\sum\limits_{i=1}^{n^{t}}\sum\limits_{i'=1}^{n^{t'}}f_i^{t}\hat{\omega}_{ii'}^{tt'}f_{i'}^{t'} \\
    =&-2\sum\limits_{t,t'=1}^T \Lambda^{tt'}{\vf^{t}}^\trans\hat{\mW}^{tt'}\vf^{t'}=-2\vf^\trans\hat{\mW}\vf \\
\end{align*}
Therefore
\begin{equation*}
    \eqref{eq:opti_pb_2} \Leftrightarrow \min\limits_{\mathbf{f}^{1},\ldots,\mathbf{f}^{T}} -2\vf^\trans\hat{\mW}\vf + 2\alpha \left(\|\vf_\ell-\vy_\ell\|^2 + \|\vf_u\|^2\right) \Leftrightarrow \eqref{eq:opti_pb_3}
\end{equation*}
The optimization problem we need to solve is
\begin{equation*}
    \min\limits_{\vf \in \mathbb{R}^{n}} \alpha\|\vf-\vy\|^2 - {\vf}^{\trans} \tilde{\mW} \vf \Leftrightarrow \min\limits_{\vf \in \mathbb{R}^{n}} {\vf}^{\trans}\left(\alpha \mI_{n}-\tilde{\mW}\right)\vf - 2\alpha \vf^{\trans} \vy  \\
\end{equation*}
The problem is convex as long as $\alpha > \|\tilde{\mW}\|$, ensuring $\alpha \mI_{n}-\tilde{\mW}$ to be positive definite. We assume that this condition is satisfied. The quantity to minimize is a quadratic form of $\vf$, so the solution of the optimization problem satisfies:
\begin{align*}
    &\min\limits_{\vf \in \mathbb{R}^{n}} \alpha\|\vf-\vy\|^2 - {\vf}^{\trans} \tilde{\mW} \vf \Leftrightarrow \left(\alpha \mI_n-\tilde{\mW}\right)\vf - \alpha\vy = 0 \\
    &\Leftrightarrow \vf = \left(\mI_{n}-\frac{\tilde{\mW}}{\alpha}\right)^{-1}\vy \Leftrightarrow \vf = \left(\mI_{n} - \frac{{\mZ}^{\trans} \mA \mZ}{Tp} \right)^{-1} \vy \\
\end{align*}
Indeed,
\begin{align*}
    \frac{\tilde{\mW}}{\alpha} &= \sum_{t,t'=1}^T \tilde{\Lambda}^{tt'}\mE_{tt'} \otimes \hat{\mW}^{tt'} = \frac{1}{Tp}\sum_{t,t'=1}^T \tilde{\Lambda}^{tt'}\mE_{tt'} \otimes ({\mX^t}^\trans\mX^{t'}) \\
    &= \frac{1}{Tp}\sum_{t,t'=1}^T (\mE_{tt} \otimes {\mX^t})^\trans(\tilde{\mLambda} \otimes \mI_p)(\mE_{t't'} \otimes \mX^{t'}) = \frac{{\mZ}^{\trans} \mA \mZ}{Tp}
\end{align*}

\section{Proof of Proposition \ref{th:optimal}} 
\label{app:optimal}
$\zeta^t\in[m_1^t,m_2^t]$ so there exists $\lambda\in[0,1]$ such that:
\begin{equation*}
    \zeta^t=\lambda m_1^t+(1-\lambda)m_2^t
\end{equation*}
\begin{align*}
    \epsilon_1^t &= \mathcal{Q}\left(\frac{\zeta^t-m_1^t}{\sigma^t}\right) = \mathcal{Q}\left((1-\lambda)\frac{m_2^t-m_1^t}{\sigma^t}\right) \\
    \epsilon_2^t &= \mathcal{Q}\left(\frac{m_2^t-\zeta^t}{\sigma^t}\right) = \mathcal{Q}\left(\lambda\frac{m_2^t-m_1^t}{\sigma^t}\right)
\end{align*}
$\mathcal{Q}$ is a decreasing function, so in order to minimize $\epsilon_1^t$ and $\epsilon_2^t$, one needs to maximize $\frac{m_2^t-m_1^t}{\sigma^t}$, or equivalently the following quantity:
\begin{equation}
    \left(\frac{m_2^{t}-m_1^t}{\sigma^{t}}\right)^2 = \frac{(\tilde{\vy}^\trans (\va_2^t-\va_1^t))^2}{\tilde{\vy}^\trans \mB^t \tilde{\vy}}
\end{equation}

Then the optimal scores rewrite (up to a multiplicative constant) as
\begin{equation*}
    \argmax_{\tilde{\vy}} \frac{(\tilde{\vy}^\trans (\va_2^t-\va_1^t))^2}{\tilde{\vy}^\trans \mB^t \tilde{\vy}} = (\mB^t)^{-1} (\va_2^t-\va_1^t)
\end{equation*}

\section{Proof of Proposition \ref{th:error}} 
\label{app:error}
To minimize $\epsilon^t=\frac{\epsilon_1^t+\epsilon_2^t}{2}$, the optimal threshold is $\zeta^t=\frac{m_1^t+m_2^t}{2}$.
\begin{equation*}
    \epsilon^t = \frac{1}{2}\mathcal{Q}\left(\frac{\zeta^t-m_1^t}{\sigma^t}\right) + \frac{1}{2}\mathcal{Q}\left(\frac{m_2^t-\zeta^t}{\sigma^t}\right) = \mathcal{Q}\left(\frac{m_2^t-m_1^t}{2\sigma^t}\right)
\end{equation*}
As stated in Proposition \ref{th:optimal}, the minimum value of $\epsilon^t$ is achieved with $\tilde{\vy}^\star=(\mB^t)^{-1}(a_2^t-a_1^t)$, which leads to
\begin{align*}
    m_2^t-&m_1^t = (a_2^t-a_1^t)^\trans \mB^t)^{-1}(a_2^t-a_1^t) \\
    \sigma^t& = \sqrt{(a_2^t-a_1^t)^\trans(\mB^t)^{-1}(a_2^t-a_1^t)}
\end{align*}
Finally, we have
\begin{equation*}
    \epsilon_\star^t = \mathcal{Q}\left(\frac{1}{2}\sqrt{(a_2^t-a_1^t)^\trans(\mB^t)^{-1}(a_2^t-a_1^t)}\right)
\end{equation*}

\section{Estimation of $\mM^\trans\mM$}
\label{app:estimation}

To estimate the matrix $\mcM = \mM^\trans\mM$, one only needs to estimate quantities such as $\theta=\vmu_1^\trans\vmu_2$. To estimate these quantities, we use the following unbiaised estimator :
\begin{equation*}
    \hat{\theta}=\frac{1}{n_1 n_2} \mathbb{1}_{n_1}^\trans \mX_1^\trans \mX_2 \mathbb{1}_{n_2}
\end{equation*}
where $(\mX_j)_{.,i} \sim \mathcal{N}(\vmu_j,\mI_p)$
\begin{equation*}
    \hat{\theta}=\frac{1}{n_1 n_2} \mathbb{1}_{n_1}^\trans \left(\mathbb{1}_{n_1}\vmu_1^\trans+\mV_1^\trans\right) \left(\vmu_2\mathbb{1}_{n_2}^\trans+\mV_2\right) \mathbb{1}_{n_2} 
\end{equation*}
where $(\mV_j)_{.,i} \sim \mathcal{N}(0,\mI_p)$ \\
To compute the law of $\hat{\theta}$, we will use the two following facts:
\begin{lemma}
    \begin{equation*}
        \frac{1}{\sqrt{n_j}}\mV_j\mathbb{1}_{n_j} \sim \mathcal{N}(0,\mI_p)
    \end{equation*}
\end{lemma}
\begin{lemma}
    If $\vu,\vv \sim \mathcal{N}(0,\mI_p)$ and are independant, then
    \begin{equation*}
        \frac{1}{\sqrt{p}}\vu^\trans \vv \sim \mathcal{N}(0,1)
    \end{equation*}
\end{lemma}

\begin{align*}
    \hat{\theta}&=\frac{1}{n_1 n_2} \mathbb{1}_{n_1}^\trans \left(\mathbb{1}_{n_1}\vmu_1^\trans+\mV_1^\trans\right) \left(\vmu_2\mathbb{1}_{n_2}^\trans+\mV_2\right) \mathbb{1}_{n_2} \\
    &=\left(\vmu_1+\frac{1}{n_1}\mV_1\mathbb{1}_{n_1}\right)^\trans \left(\vmu_2+\frac{1}{n_2}\mV_2\mathbb{1}_{n_2}\right) \\
    &=\left(\vmu_1+\frac{1}{\sqrt{n_1}}\vu_1\right)^\trans \left(\vmu_2+\frac{1}{\sqrt{n_2}}\vu_2\right) \\
    &\text{with} \quad \vu_1,\vu_2 \sim\mathcal{N}(0,\mI_p) \\
    &=\vmu_1^\trans\vmu_2 + \frac{1}{\sqrt{n_2}}\vmu_1^\trans\vu_2 + \frac{1}{\sqrt{n_1}}\vu_1^\trans\vmu_2 + \frac{1}{\sqrt{n_1 n_2}} \vu_1^\trans\vu_2 \\
    &=\vmu_1^\trans\vmu_2 + \frac{\|\vmu_1\|}{\sqrt{n_2}}z_1 + \frac{\|\vmu_2\|}{\sqrt{n_1}}z_2 + \sqrt{\frac{p}{n_1 n_2}} z_3 \\
    &\text{with} \quad z_1,z_2,z_3 \sim\mathcal{N}(0,1).
\end{align*}
Therefore the estimator is consistent, and converges with an $\mathcal{O}\left(\frac{1}{\sqrt{n}}\right)$ speed, under Assumption \ref{ass:growth_rate}. For the specific case of $\theta=\vmu_1^\trans\vmu_1$, one can use the previous results, but needs to divide the samples in two separate subsets, in order to keep the property of independance between the samples.

\section{Proof of Theorem \ref{th:main_bis}}

The proof is organized as follows:
\begin{itemize}
    \item A first order deterministic equivalent (Section \ref{app:order_1_calculus}) is needed to compute the mean of $f_i$ (Section \ref{app:mean_calculus})
    \item A second order deterministic equivalent (Section \ref{app:order_2_calculus}) is needed to compute the variance of $f_i$ (Section \ref{app:variance_calculus})
\end{itemize}

A more detailed version of the proof is available at \href{https://gricad-gitlab.univ-grenoble-alpes.fr/legervi/tsp}{https://gricad-gitlab.univ-grenoble-alpes.fr/legervi/tsp}.

In the following proof, we will need to distinguish the data matrix $\mZ$ and its centered version $\mathring{\mZ}$, as the centering breaks the independance between samples. This effect of centering is sumed up in the following lemma. \\
\begin{lemma}
    For a data matrix $\mX$ defined as in Assumption \ref{ass:data_distribution_2}, and centered as in equation \eqref{eq:centering}, we have asymtotically:
    \begin{equation}
    \label{eq:xxt}
        \mathbb{E}\left[\mathring{\mX}\mathring{\mX}^\trans\right] = (1-\frac{1}{n})\mathbb{E}\left[\mX\mX^\trans\right] - \frac{1}{n}\mathbb{E}[\mX]\mathbb{E}[\mX]^\trans
    \end{equation}
    \begin{equation}
    \label{eq:xtx}
        \left\{
            \begin{array}{lr}
                \mathbb{E}\left[\mX^\trans\mX\right] = \mathbb{E}[\mX]^\trans\mathbb{E}[\mX] + \tr(\mSigma)\mI_n, \\
                \mathbb{E}\left[\mathring{\mX}^\trans\mathring{\mX}\right] = \mathbb{E}[\mathring{\mX}]^\trans\mathbb{E}[\mathring{\mX}] + \tr(\mSigma)\mI_n - \frac{1}{n}\tr(\mSigma)\mathbb{1}_n\mathbb{1}_n^\trans
            \end{array}
        \right.
    \end{equation}
\end{lemma}
While the matrix $\mX\mX^\trans$ is practically unaffected by the centering, the matrix $\mX^\trans\mX$ is changed, not only by the centering of the mean matrix $\mathbb{E}[\mX]^\trans\mathbb{E}[\mX]$, but also with the appearance of the bias term $-\frac{1}{n}\tr(\mSigma)\mathbb{1}\mathbb{1}^\trans$. With these results in mind, $\mX^t,\vmu_j^t,\mZ$ will still denote their centered versions, as it was the case until now.

\subsection{First order deterministic equivalent}
\label{app:order_1_calculus}

We introduce the notation $\mF\leftrightarrow \bar{\mF}$ which denote that $\bar{\mF}$ is a deterministic equivalent of $\mF$. \\
A deterministic equivalent, say $\bar{\mF}\in\mathbb{R}^{n\times p}$, of a given random matrix $\mF\in\mathbb{R}^{n\times p}$, is defined by the fact that under assumption \ref{ass:growth_rate}, for any deterministic linear functional $f:\mathbb{R}^{n\times p}\to\mathbb{R}$, $f(\mF-\bar{\mF})\to 0$ almost surely (for instance, for $\vu,\vv$ of unit norm, $\vu^\trans (\mF-\bar{\mF})\vv\asto 0$ and, for $\mA\in\mathbb{R}^{p\times n}$ deterministic of bounded operator norm, $\frac{1}{n}\tr \mA(\mF-\bar{\mF})\asto 0$).

Theorem $2.8$ of \cite{couillet_liao_2022} provides us a deterministic equivalent for $\mQ$. To express it, we need to define

\begin{equation}
    \mC_{j}^{t}=\mA^{\frac 12}\left(\mE_{tt}\otimes \left(\mathbf{\Sigma}_j^{t}+\boldsymbol{\mu}_j^{t}{\boldsymbol{\mu}_j^{t}}^{\trans}\right)\right)\mA^{\frac 12}
\end{equation}

Then, we have $\mQ \leftrightarrow \bar{\mQ} = \left(\mI_{Tp} - \sum_{t,j} \tilde{\delta}_j^t \mC_j^t\right)^{-1}$, with $\tilde{\delta}_j^t$ satisfying the following system of equations:

\begin{equation*}
    \tilde{\delta}_j^t = \frac{\rho_{j}^t}{Tc}\frac{1}{1-\delta_j^t} \quad \text{and} \quad
    \delta_j^t = \frac{1}{Tp}\tr \left(\mC_j^t \bar{\mQ} \right)
\end{equation*}

Let's compute more explicitly the quantity $\delta_j^t$. If the covariances of the data for all class $j$ and all tasks $t$ are isotropic of same magnitude, i.e  $\mathbf{\Sigma}_{j}^{t}=\mI_p$, then, as $\boldsymbol{\mu}_j^{t}{\boldsymbol{\mu}_j^{t}}^{\trans}$ is a small rank perturbation of $\mathbf{\Sigma}_{j}^{t}$, we have asymptotically:
\begin{equation*}
    \tr \left(\mC_j^{t}\bar{\mQ}\right) = \tr\left(\mA^{\frac 12}\left(\mE_{tt}\otimes \mI_p\right)\mA^{\frac 12}\bar{\mQ}\right)
\end{equation*}
Therefore, $\delta_1^t=\delta_2^t$, and we will use the same notation $\delta^t$ for both in the following. Let's define $\bar{\delta}^{t}=\tilde{\delta}_1^{t}+\tilde{\delta}_2^{t}$, and $\bar{\mM}=[\ve_1^{[T]}\otimes \boldsymbol{\mu}_1^{1},\ve_1^{[T]}\otimes \boldsymbol{\mu}_2^{1},\ldots,\ve_T^{[T]}\otimes \boldsymbol{\mu}_2^{T}]$. Then

\begin{align*}
    &\sum_{t,j} \tilde{\delta}_j^t \mC_j^t = \sum_{t,j} \tilde{\delta}_j^t \mA^{\frac 12}\left(\mE_{tt}\otimes \left(\mI_p+\boldsymbol{\mu}_j^{t}{\boldsymbol{\mu}_j^{t}}^{\trans}\right)\right)\mA^{\frac 12} \\
    =&\left(\tilde{\mathbf{\Lambda}}^{\frac 12}\mathcal{D}_{\bar{\boldsymbol{\delta}}}\tilde{\mathbf{\Lambda}}^{\frac 12}\right)\otimes \mI_p + \mA^{\frac 12} \bar{\mM}\mathcal{D}_{\tilde{\boldsymbol{\delta}}}\bar{\mM}^\trans \mA^{\frac 12}
\end{align*}
If $\bar{\mQ}_0=\left(\mI_T-\tilde{\mathbf{\Lambda}}^{\frac 12}\mathcal{D}_{\bar{\boldsymbol{\delta}}}\tilde{\mathbf{\Lambda}}^{\frac 12}\right)^{-1}\otimes \mI_p$, using Woodbury identity,
\begin{align*}
    \bar{\mQ}
 &=\bar{\mQ}_0 + \bar{\mQ}_0\mU\left(\mI_{2T}-\mU^\trans\bar{\mQ}_0\mU\right)^{-1}\mU^\trans\bar{\mQ}_0
\end{align*}
The second term is a matrix of rank $2T<<Tp$ and can be neglected in the trace. Then

\begin{align*}
    \mA^{\frac 12}\bar{\mQ}_0 \mA^{\frac 12} 
    &= \underbrace{\left[\tilde{\mathbf{\Lambda}} + \tilde{\mathbf{\Lambda}}\left(\mathcal{D}_{\bar{\boldsymbol{\delta}}}^{-1}-\tilde{\mathbf{\Lambda}}\right)^{-1}\tilde{\mathbf{\Lambda}}\right]}_{=\mcA} \otimes \mI_p.
\end{align*}
Finally,
\begin{equation*}
    \delta^{t}=\frac{1}{Tp}\tr \left(\mE_{tt}\mcA\right)\tr(\mI_p)=\frac{1}{T}\mcA_{tt}.
\end{equation*}

We can then conclude that the $\{\delta^{t}\}_{t}$ are the solution of the following system of equations:

\begin{equation*}
    \left\{
  \begin{array}{lr}
    \forall t, ~ \delta^{t}=\frac{1}{T}\mcA_{tt},\\
    \mcA=\tilde{\mathbf{\Lambda}} + \tilde{\mathbf{\Lambda}}\left(\diag_{\bar{\boldsymbol{\delta}}}^{-1}-\tilde{\mathbf{\Lambda}}\right)^{-1}\tilde{\mathbf{\Lambda}}
  \end{array}
\right.
\end{equation*}

\subsection{Computation of the mean}
\label{app:mean_calculus}

\begin{equation*}
    \mathbb{E}[f_i] = \frac{1}{Tp} \sum_{i'} \mathbb{E}\left[\vz_i^{\trans}\mA^{\frac 12} \mQ \mA^{\frac 12}\vz_{i'}\right] y_{i'}.
\end{equation*}

Using Sherman-Morrison formula, it can be proved that:
\begin{equation}
\label{eq:sherman1}
    \mQ = \mQ_{-i} + \frac{1}{Tp}\frac{\mQ_{-i}\mA^{\frac 12}\vz_i\vz_i^{\trans}\mA^{\frac 12}\mQ_{-i}}{1-\frac{1}{Tp}\vz_i^{\trans}\mA^{\frac 12}\mQ_{-i}\mA^{\frac 12}\vz_i}
\end{equation}
with $\mQ_{-i}=\left(\mI_{Tp}-\frac{\mA^{\frac 12}\mZ_{-i}\mZ_{-i}^\trans \mA^{\frac 12}}{Tp}\right)^{-1}$, the notation $\mZ_{-i}$ standing for the matrix $\mZ$ with the $i$-th column removed.

In particular:
\begin{align*}
   &\mathbb{E}\left[\vz_i^{\trans}\mA^{\frac 12}\mQ\mA^{\frac 12}\vz_{i'}\right] = \frac{\mathbb{E}\left[\vz_i^{\trans}\mA^{\frac 12}\mQ_{-i}\mA^{\frac 12}\vz_{i'}\right]}{1-\delta^{t}}.
\end{align*}


And thanks to equation \eqref{eq:xtx}
\begin{align*}
    &\mathbb{E}\left[\vz_i^{\trans}\mA^{\frac 12}\mQ_{-i,-i'}\mA^{\frac 12}\vz_{i'}\right] \\
    &=\mathbb{E}\left[\vz_i\right]^{\trans}\mA^{\frac 12}\bar{\mQ}\mA^{\frac 12}\mathbb{E}\left[\vz_{i'}\right] - \frac{1}{n^t}\mathbb{E}\left[\tr(\mC^t\mQ)\right]\mathbb{1}_{t=t'} \\
    &={\ve_{t,j}^{[2T]}}^\trans\bar{\mM}^\trans\mA^{\frac 12}\bar{\mQ}\mA^{\frac 12}\bar{\mM}\ve_{t',j'}^{[2T]} - \frac{Tp}{n^t}\delta^t{\ve_{t,j}^{[2T]}}^\trans(\mI_T \otimes \mathbb{1}_2\mathbb{1}_2^\trans)\ve_{t',j'}^{[2T]},
\end{align*}
with
\begin{equation*}
    \mC^{t} := \mA^{\frac 12}\left(\mE_{tt}\otimes \mI_p\right)\mA^{\frac 12}.
\end{equation*}
So finally
\begin{align*}
    &\mathbb{E}\left[\vz_i^{\trans}\mA^{\frac 12}\mQ\mA^{\frac 12}\vz_{i'}\right] \simeq \frac{\mathbb{E}\left[\vz_i^{\trans}\mA^{\frac 12}\mQ_{-i,-i'}\mA^{\frac 12}\vz_{i'}\right]}{(1-\delta^t)(1-\delta^{t'})} \\
    &=\frac{{\ve_{t,j}^{[2T]}}^\trans}{1-\delta^t}\left(\bar{\mM}^\trans\mA^{\frac 12}\bar{\mQ}\mA^{\frac 12}\bar{\mM} - \frac{Tc}{\rho^t}\delta^t \mI_T \otimes \mathbb{1}_2\mathbb{1}_2^\trans\right)\frac{\ve_{t',j'}^{[2T]}}{1-\delta^{t'}}.
\end{align*}
Going back to the quantity we want to estimate,

\begin{align*}
    &\sum_{i'} \frac{y_{i'}}{Tp} \frac{\ve_{t',j'}^{[2T]}}{1-\delta^{t'}} 
    = \diag_{\tilde{\vdelta}} \diag_{\veta}\bar{\mD} \tilde{\vy}
\end{align*}
with $\bar{d}_{j_1,j_2}^t = \frac{1}{n_{\ell j_1}^t}\sum_{i'|x_{i}\in \mathcal{C}_{j_1}^{t}} d_{i'j_2}^t$ and 
\begin{equation*}
    \bar{\mD}=\sum_{t=1}^T \mE_{tt}\otimes \begin{pmatrix} \bar{d}_{11}^t & \bar{d}_{12}^t\\ \bar{d}_{21}^t & \bar{d}_{22}^t \end{pmatrix},
\end{equation*}
so we have 
\begin{equation*}
    \mathbb{E}[f_i] = \frac{{\ve_{t,j}^{[2T]}}^\trans}{1-\delta^t}\left(\bar{\mM}^\trans\mA^{\frac 12}\bar{\mQ}\mA^{\frac 12}\bar{\mM} - \frac{Tc}{\rho^t}\delta^t\mI_T \otimes \mathbb{1}_2\mathbb{1}_2^\trans\right)\diag_{\tilde{\vdelta}} \diag_{\veta}\bar{\mD} \tilde{\vy}.
\end{equation*}
Let us define 
\begin{equation*}
    \mTheta := \bar{\mM}^\trans\mA^{\frac 12}\bar{\mQ}\mA^{\frac 12}\bar{\mM} = \mTheta_0\left(\mI_{2T}-\diag_{\tilde{\vdelta}}\mTheta_0\right)^{-1}
\end{equation*}
\begin{equation*}
    \mathbb{E}[f_i] = \frac{{\ve_{t,j}^{[2T]}}^\trans}{1-\delta^t}\left(\mTheta - \frac{Tc}{\rho^t}\delta^t\mI_T \otimes \mathbb{1}_2\mathbb{1}_2^\trans\right)\diag_{\tilde{\vdelta}} \diag_{\veta} \bar{\mD} \tilde{\vy}
\end{equation*}
As $\mGamma=\mI_T \otimes \mathbb{1}_2\mathbb{1}_2^\trans$ and $\gamma^t=\frac{Tc\delta^t}{\rho^t}$, we can conclude that:

\begin{equation}
    m_j^t = \mathbb{E}[f_i] = \frac{{\ve_{t,j}^{[2T]}}^\trans}{1-\delta^t}\left(\mTheta-\gamma^t\mGamma\right) \diag_{\tilde{\vdelta}} \diag_{\veta}\bar{\mD}\tilde{\vy}
\end{equation}

\subsection{Second order deterministic equivalent}
\label{app:order_2_calculus}

In the following, we will need a deterministic equivalent for $\mQ\mC_j^{t}\mQ$ (which is \textit{not} $\bar{\mQ}\mC_j^{t}\bar{\mQ}$). An analysis similar to the one performed in \ref{app:order_1_calculus}, allows to show that :

\begin{align*}
    & \mathbb{E}\left[\vu^\trans \mQ\mC_j^{t}\mQ\vv\right] - \vu^\trans \bar{\mQ}\mC_j^{t}\bar{\mQ}\vv \\
    &=\sum_{t',j'} \frac{n_{j'}^{t'}}{(Tp)^2(1-\delta^{t'})^2}\tr \left(\bar{\mQ}\mC_j^{t} \bar{\mQ}\mC_{j'}^{t'}\right) \mathbb{E}\left[\vu^\trans\mQ\mC_{j'}^{t'}\mQ\vv\right]
\end{align*}
Until now we have the following (not deterministic) equivalent:
\begin{equation}
\label{eq:eq_non_deterministic}
    \mQ\mC_j^{t}\mQ\leftrightarrow \bar{\mQ}\mC_j^{t}\bar{\mQ}+\sum_{t',j'} \frac{n_{j'}^{t'}}{(Tp)^2(1-\delta^{t'})^2}\tr \left(\bar{\mQ}\mC_j^{t} \bar{\mQ} \mC_{j'}^{t'}\right) \mQ\mC_{j'}^{t'}\mQ
\end{equation}

First of all, as $\mC_j^{t}$ is a small-rank perturbation of $\mC^{t}$ and $\bar{\mQ}$ is a small-rank perturbation of $\bar{\mQ}_0$:
\begin{align*}
    &\frac{1}{Tp}\tr \left(\bar{\mQ}\mC_j^{t}\bar{\mQ}\mC_{j'}^{t'}\right) = \frac{1}{Tp}\tr \left(\bar{\mQ}_0\mC^{t}\bar{\mQ}_0\mC^{t'}\right) \\
    &=\frac{1}{T}\tr \left(\mE_{tt}\mcA\mE_{t't'}\mcA\right) = \frac{\mcA_{tt'}^2}{T}
\end{align*}
Let's define 
\begin{align*}
    \bar{\mS}^{tt'}&=\frac{1}{Tc(1-\delta^{t'})^2}\frac{1}{Tp}\tr \left(\bar{\mQ}\mC^{t}\bar{\mQ}\mC^{t'}\right) = \frac{\mcA_{tt'}^2}{T^2c(1-\delta^{t'})^2} \\
    \mS^{tt'}&=\frac{1}{Tc(1-\delta^{t'})^2}\frac{1}{Tp}\mathbb{E}\left[\tr \left(\mQ\mC^{t}\mQ\mC^{t'}\right)\right]
\end{align*}

One can deduce from \eqref{eq:eq_non_deterministic} that $\mS = \bar{\mS} + \bar{\mS}\diag_{\bar{\vrho}}\mS$, so $\mS = \bar{\mS}\left(\mI_T-\diag_{\bar{\vrho}}\bar{\mS}\right)^{-1}$. Similarly, it follows that:
\begin{equation}
    \mQ\mC_j^{t}\mQ \leftrightarrow \bar{\mQ}\mC_j^{t}\bar{\mQ} + \sum_{t',j'} \rho_{j'}^{t'}\mS^{tt'} \bar{\mQ}\mC_{j'}^{t'}\bar{\mQ}
\end{equation}

\subsection{Computation of the variance}
\label{app:variance_calculus}
Under assumption \ref{ass:growth_rate}, the second order moment can be computed as:

\begin{align*}
    &\mathbf{E}[f_i^2] 
    =\underbrace{\frac{1}{(Tp)^2}\sum_{i'} \mathbb{E}\left[\vz_{i'}^\trans \mA^{\frac 12}\mQ\mA^{\frac 12}\vz_i\vz_i^\trans \mA^{\frac 12}\mQ\mA^{\frac 12}\vz_{i'}\right] y_{i'}^2}_{:=C_1} \\
    &+\underbrace{\frac{1}{(Tp)^2}\sum_{i'\neq i''} \mathbb{E}\left[\vz_{i'}^\trans \mA^{\frac 12}\mQ\mA^{\frac 12}\vz_i\vz_i^\trans \mA^{\frac 12}\mQ\mA^{\frac 12}\vz_{i''}\right] y_{i'} y_{i''}}_{:=C_2}
\end{align*}

\begin{itemize}
    \item Computation of $C_1$
    \begin{align*}
        &C_1=\frac{1}{(Tp)^2}\sum_{i'} \frac{\mathbb{E}\left[\tr\left(\mQ\mC^t\mQ\mC^{t'}\right)\right]}{(1-\delta^t)^2(1-\delta^{t'})^2}y_{i'}^2 \\
        &=\sum_{t'} \frac{\rho^{t'}\mS^{tt'}}{(1-\delta^t)^2}\eta^{t'}\begin{pmatrix}
    \tilde{y}_{1}^{t'} & \tilde{y}_{2}^{t'} \end{pmatrix}\begin{pmatrix}
    \tilde{d}_{11}^{t'} & \tilde{d}_{12}^{t'} \\
    \tilde{d}_{21}^{t'} & \tilde{d}_{22}^{t'} \end{pmatrix}\begin{pmatrix}
    \tilde{y}_{1}^{t'} \\
    \tilde{y}_{2}^{t'} \end{pmatrix} \\
    &=\tilde{\vy}^\trans\left(\mT^t\odot\tilde{\mD}\right)\tilde{\vy},
    \end{align*}
     with $\tilde{d}_{j_1j_2}^t = \frac{1}{n_\ell^t}\sum_{i|\vx_{i}\in\mathcal{C}^{t}} d_{ij_1}^t d_{ij_2}^t$, if we further define 
    \begin{align*}
        \tilde{\mD}&=\sum_{t=1}^T \mE_{tt}\otimes \begin{pmatrix}
            \tilde{d}_{11}^t & \tilde{d}_{12}^t\\
            \tilde{d}_{21}^t & \tilde{d}_{22}^t \end{pmatrix} \\
        \mT^t&=\frac{1}{(1-\delta^t)^2}\diag_{\bar{\vrho}\odot\bar{\veta}\odot\mS^{t.}}\otimes\mathbb{1}_2\mathbb{1}_2^\trans.
    \end{align*}
   
    \item Computation of $C_2$ \\
    Using \ref{eq:sherman1} and \ref{eq:xtx}, we have
    \begin{align*}
        &\mathbb{E}\left[\vz_{i'}^\trans \mA^{\frac 12}\mQ\mA^{\frac 12}\vz_i\vz_i^\trans \mA^{\frac 12}\mQ\mA^{\frac 12}\vz_{i''}\right] \\
        &=\frac{{\ve_{t',j'}^{[2T]}}^\trans}{1-\delta^{t'}}\underbrace{\frac{\bar{\mM}^\trans \mA^{\frac 12}\mathbb{E}\left[\mQ\mC_j^t\mQ\right]\mA^{\frac 12}\bar{\mM}}{(1-\delta^t)^2}}_{:=\mathbf{\mcU}_1^t}\frac{\ve_{t'',j''}^{[2T]}}{1-\delta^{t''}} \\
        &-Tc\mS^{tt'}{\ve_{t',j'}^{[2T]}}^\trans\underbrace{\frac{\mGamma}{(1-\delta^t)^2}}_{:=\mathbf{\mcU}_2^t}\frac{Tc}{\rho^{t'}}\ve_{t'',j''}^{[2T]} \\
        &+Tc{\ve_{t',j'}^{[2T]}}^\trans\left(\underbrace{\frac{(\mTheta-\gamma^t\mGamma)}{(1-\delta^t)^2}}_{:=\mathbf{\mcU}_3^t}\frac{\mS^{tt'}}{1-\delta^{t''}}+\frac{\mS^{tt''}}{1-\delta^{t'}}\mathbf{\mcU}_3^t\right)\ve_{t'',j''}^{[2T]}. 
    \end{align*}
    Let us decompose 
    \begin{align*}
        &\bar{\mM}^\trans \mA^{\frac 12} \bar{\mQ}_0\left(\mC_j^t + \sum_{t',j'}\rho_{j'}^{t'}\mS^{tt'}\mC_{j'}^{t'}\right)\bar{\mQ}_0 \mA^{\frac 12} \bar{\mM} \\
        &=\mTheta_0\left(\ve_{t,j}^{[2T]}{\ve_{t,j}^{[2T]}}^\trans+\diag_{\vr^t}\right)\mTheta_0 + \underbrace{\left(\bar{\mcV}^t \otimes \mathbb{1}_2\mathbb{1}_2^\trans\right) \odot \mM^\trans\mM}_{:=\bar{\mOmega}_0^t},
    \end{align*}
    with $\vr^t = \vrho\odot\left(\mS^{t}\otimes\mathbb{1}_2\mathbb{1}_2^\trans\right)$ and $\bar{\mcV}^t = \mcV^t + \sum_{t'}\rho^{t'}\mS^{tt'}\mcV^{t'}$. We thus have:
    \begin{align*}
        &\bar{\mM}^\trans \mA^{\frac 12} \mathbb{E}[\mQ\mC_j^t\mQ]\mA^{\frac 12} \bar{\mM} \\
        &=\mTheta \left(\ve_{t,j}^{[2T]}{\ve_{t,j}^{[2T]}}^\trans+\diag_{\vr^t}\right)\mTheta \\
        &+\underbrace{\left(\mI_{2T}-\mTheta_0\diag_{\tilde{\vdelta}}\right)^{-1}\bar{\mOmega}_0^t\left(\mI_{2T}-\diag_{\tilde{\vdelta}}\mTheta_0\right)^{-1}}_{:=\bar{\mOmega}^t}.
    \end{align*}
    \begin{align*}
        C_2&=\frac{1}{(1-\delta^t)^2}\tilde{\vy}^\trans\bar{\mD}^\trans\diag_{\veta}\left[\diag_{\tilde{\vdelta}}\mTheta \left(\ve_{t,j}^{[2T]}{\ve_{t,j}^{[2T]}}^\trans+\diag_{\vr^t}\right)\mTheta\diag_{\tilde{\vdelta}} \right. \\
        &+\diag_{\tilde{\vdelta}}\bar{\mOmega}^t\diag_{\tilde{\vdelta}} - \mGamma^t\diag_{\vr^t} + 2\diag_{\tilde{\vdelta}}(\mTheta-\gamma^t\mGamma)\diag_{\vr^t}\left.\right]\diag_{\veta}\bar{\mD} \tilde{\vy} 
    \end{align*}
    
\end{itemize}
As $Var(q_i)=C_1+C_2-{m_j^t}^2$, we finally have:
\begin{align*}
    &{\sigma^t}^2 = \frac{1}{(1-\delta^t)^2}\tilde{\vy}^\trans\left(\mT^t\odot\tilde{\mD}\right)\tilde{\vy} \\
    &+\frac{1}{(1-\delta^t)^2}\tilde{\vy}^\trans\bar{\mD}^\trans\diag_{\veta}\diag_{\tilde{\vdelta}}\left(\mTheta\diag_{\vr^t}\mTheta+\bar{\mOmega}^t-{\gamma^t}^2\mGamma^t\right)\diag_{\tilde{\vdelta}} \diag_{\veta}\bar{\mD} \tilde{\vy} \\
    &+\frac{1}{(1-\delta^t)^2}\tilde{\vy}^\trans\bar{\mD}^\trans\diag_{\veta}\left[2\diag_{\tilde{\vdelta}}(\mTheta-\gamma^t\mGamma) - \mGamma^t\right] \diag_{\vr^t}\diag_{\veta}\bar{\mD} \tilde{\vy},
\end{align*}
with
\begin{equation*}
    \mathbf{\mcU}=\mTheta\ve_{t,j}^{[2T]}{\ve_{t,j}^{[2T]}}^\trans\mTheta - {\gamma^t}^2\underbrace{\mE^{tt} \otimes (\mathbb{1}_2\mathbb{1}_2^\trans)}_{:=\mGamma^t}.
\end{equation*}

\bibliographystyle{IEEEtran}
\bibliography{mybibfile}

\vfill

\end{document}

%% file: math_command.tex
\def\R{\mathbb R}

\def\P{\mathbb P}

\def\Cov{\mathrm{Cov}}
\renewcommand\epsilon{\varepsilon}

\def\va{{\mathbf{a}}}

\def\ve{{\mathbf{e}}}
\def\vf{{\mathbf{f}}}

\def\vr{{\mathbf{r}}}

\def\vu{{\mathbf{u}}}
\def\vv{{\mathbf{v}}}

\def\vx{{\mathbf{x}}}

\def\vy{{\mathbf{y}}}
\def\vz{{\mathbf{z}}}

\def \vmu{{\boldsymbol{\mu}}}

\def \vrho{{\boldsymbol{\rho}}} 
\def \veta{{\boldsymbol{\eta}}} 
\def \vdelta{{\boldsymbol{\delta}}}

\def\mA{{\mathbf{A}}}
\def\mB{{\mathbf{B}}}
\def\mC{{\mathbf{C}}}
\def\mD{{\mathbf{D}}}
\def\mE{{\mathbf{E}}}
\def\mF{{\mathbf{F}}}

\def\mI{{\mathbf{I}}}

\def\mM{{\mathbf{M}}}

\def\mP{{\mathbf{P}}}
\def\mQ{{\mathbf{Q}}}

\def\mS{{\mathbf{S}}}
\def\mT{{\mathbf{T}}}
\def\mU{{\mathbf{U}}}
\def\mV{{\mathbf{V}}}
\def\mW{{\mathbf{W}}}
\def\mX{{\mathbf{X}}}

\def\mZ{{\mathbf{Z}}}
\def\mSigma{{\mathbf{\Sigma}}}
\def\mGamma{{\mathbf{\Gamma}}}
\def\mLambda{{\mathbf{\Lambda}}} 
\def\mTheta {{\mathbf{\Theta }}}
\def\mOmega {{\mathbf{\Omega }}}

\def\mcA{{\boldsymbol{\mathcal{A}}}}
\def\mcM{{\boldsymbol{\mathcal{M}}}}
\def\mcU{{\boldsymbol{\mathcal{U}}}}
\def\mcV{{\boldsymbol{\mathcal{V}}}}

\def\R{\mathbb{R}}
\def\P{\mathbb{P}}

\def\trans{{\sf T}}
\def\diag{{\mathcal{D}}}
\def\tr{\rm tr}
\DeclareMathOperator*{\argmax}{arg\,max}

\newcommand{\ie}{\emph{i.e.,}~}
\pgfmathdeclarefunction{gauss}{2}{%
  \pgfmathparse{1/(#2*sqrt(2*pi))*exp(-((x-#1)^2)/(2*#2^2))}%
}
\newcommand{\asto}{\overset{\rm a.s.}{\longrightarrow}}